\documentclass[fleqn,10pt]{wlscirep}

\usepackage{graphicx,epsfig}

\usepackage{times}
\usepackage{graphics,dcolumn,bm,fleqn,float}
\usepackage{amssymb,amsmath,multirow,rotate,color}
\usepackage{comment} 
\usepackage{cite} 
\usepackage{stackengine}
\usepackage[normalem]{ulem}
\DeclareMathAlphabet{\altmathcal}{OMS}{cmsy}{m}{n}

\usepackage{amsthm,amsmath}
\usepackage[utf8]{inputenc} 

\usepackage{color}
\usepackage{ulem}

\usepackage{amssymb}

\usepackage{url}

\newcommand{\specialcell}[2][c]{%
	\begin{tabular}[#1]{@{}c@{}}#2\end{tabular}}

\title{Predicting the impact of urban change in pedestrian and road safety}

\author[1]{Cristina Bustos}
\author[1]{Daniel Rhoads}
\author[1]{Agata Lapedriza}
\author[1]{Javier Borge-Holthoefer}
\author[1]{Albert Solé-Ribalta}
\affil[1]{Internet Interdisciplinary Institute (IN3), Universitat Oberta de Catalunya, Barcelona, Catalonia, Spain}

\begin{abstract} 
Increased interaction between and among pedestrians and vehicles in the crowded urban environments of today gives rise to a negative side-effect: a growth in traffic accidents, with pedestrians being the most vulnerable elements. Recent work has shown that Convolutional Neural Networks are able to accurately predict accident rates exploiting Street View imagery along urban roads. The promising results point to the plausibility of aided design of safe urban landscapes, for both pedestrians and vehicles. In this paper, by considering historical accident data and Street View images, we detail how to automatically predict the impact (increase or decrease) of urban interventions on accident incidence. The results are positive, rendering an accuracies ranging from 60 to 80\%. We additionally provide an interpretability analysis to unveil which specific categories of urban features impact accident rates positively or negatively. Considering the transportation network substrates (sidewalk and road networks) and their demand, we integrate these results to a complex network framework, to estimate the effective impact of urban change on the safety of pedestrians and vehicles. Results show that public authorities may leverage on machine learning tools to prioritize targeted interventions, since our analysis show that limited improvement is obtained with current tools. Further, our findings have a wider application range such as the design of safe urban routes for pedestrians or to the field of driver-assistance technologies.
\end{abstract}

\begin{document}

\maketitle

\section*{Introduction}
Despite significant public investment in their prevention, traffic fatalities remain a leading cause of death around the world \cite{world2018global}. Accordingly, the past several years have seen more and more cities publicly committing to the Vision Zero initiative \cite{tingvall1999vision}, a global standard originally promoted by the Swedish parliament, which aims at setting a goal of 0 yearly traffic fatalities. With vast amounts of data about cities currently available, automated analysis has been recognized as a crucial tool to help urban planners in decision tasks \cite{kandt2021smart,ibrahim2020understanding,he2021inferring, yoshimura2021street}. While the link between the growth of urban open data and new computational techniques such as Deep Learning has been pointed out by several authors \cite{naik2017computer,alhasoun2019streetify,song2018farsa,bustos2021explainable,kauer2018mapping,simini2021deep,dubey2016deep}, much work can still be done in leveraging these new technologies towards the end of achieving the ambitious goals of Vision Zero. 

A crucial step in this direction is to better understand the structural and dynamic properties of pedestrian, bike and urban networks that cause accidents. Of course, a large amount of factors impact accident incidence in urban environments, including driver \cite{wu2021influence} and pedestrian distraction \cite{nasar2008mobile}, traffic network structure \cite{rifaat2011effect,moeinaddini2014relationship,daraei2021data}, demographic variables (socio-economic status, race, gender) \cite{mukoko2019examining}, and more. Among these, built environment features, such as the organization and complexity of the visual stimuli at particular locations, are important, but relatively understudied at large scales. Many studies in transportation planning and traffic safety have analyzed the characteristics of particular urban intersections and street segments, in relation to crash data, in order to extrapolate which features contribute to accident incidence \cite{nasar2008mobile,mukoko2019examining,mecredy2012neighbourhood,fu2019investigating,hu2018dangerous}. Up to now, due to lack of available (or open) data and tools for automated analysis, these studies have mostly been constrained either to the limited features available from planimetric map data \cite{rifaat2011effect,moeinaddini2014relationship,ukkusuri2012role}, or to detailed, but time-consuming, descriptions of particular settings collected by hand \cite{mecredy2012neighbourhood,fu2019investigating}. However, the rise of new Big Data sources \cite{zhang2020prediction,dubey2016deep,chen2016effects,palazzi2018predicting} related to urban environments and transportation has boosted the development of new techniques \cite{bustos2021explainable,khaki2021probabilistic,palazzi2018predicting,sargoni2020sequential} and the combination of complementary tools existing in different fields \cite{bogacz2021modelling,he2021inferring}, such as complex systems \cite{daraei2021data,alhazzani2021urban}.

In previous work \cite{bustos2021explainable}, with the aim of clarifying the relation between pedestrian and vehicle accidents and built environment features, we compiled and analyzed a large dataset of street-level images obtained from Google Street View \cite{anguelov2010google} and the crowd-sourced Mapillary \cite{mapillary2019} collection. These datasets provide valuable fine-grained information on the location and organization of urban objects, and even dynamic information in terms of existing mobile objects (pedestrians, bicycles, cars and public transport). The Deep Learning model in  \cite{bustos2021explainable}, taking as inputs (1) driver point-of-view images, and (2) geotagged historical accident data, shows notable performance in predicting accident rates at a very high spatial resolution (approx. 15 meters). The high levels of accuracy achieved seem to indicate that the presence, absence, or particular arrangement of urban objects can indeed affect accident rates. Further, simple and well-organized scenes seem to be related to lower accident rates. The causal plausibility of this uncovered correlation can be supported by extensive literature on attention processes \cite{nasar2008mobile,moray1959attention,kahneman1973attention,alvarez2004capacity,richards2010development}, which confirm that complex scenes are harder to process (in terms of division of attention) by drivers and pedestrians, which could logically lead to an increase in driver's stress \cite{bustos2021stress} or accident rates \cite{bustos2021explainable}. 

The urban environment is constantly changing as a response to shifting mobility patterns, new construction, or government intervention. Several studies have already taken advantage of street-level imagery to automate the analysis of urban change \cite{naik2017computer,ito2021assessing,lu2019using,alhasoun2019streetify}, road safety assessment \cite{song2018farsa,kita2019google}, perceived risk \cite{naik2014streetscore}, urban demographics \cite{gebru2017pnas} and others \cite{dubey2016deep,dubey2016deep}, but much remains to be done to understand and design the physical substrate necessary for safe and efficient pedestrian and vehicle navigation networks \cite{xu2020towards,rhoads2020planning,de2015personalized,daraei2021data}. In this work, leveraging our recent advances in the prediction of vehicle and pedestrian hazard, we address the prediction of the impact urban change may have on accident rates in the city of Barcelona, Catalunya. We initially motivate our work with a detailed analysis of the evolution of accident rates and, subsequently, we show that our automated methods are able to predict, with sufficient accuracy, how such structural changes and interventions will affect accident rates. We, finally, combine our hazard estimation with network theory to put in contrast pedestrian safety with sidewalk and road demand. 

The rest of the paper is organized as follows. Section \ref{sec:methods} describes (1) the collection, curation and construction of the different datasets that are used to draw and test predictions; (2) the Deep Learning architecture employed; and (3) a brief overview of traffic flow estimation for pedestrians and vehicles. Section \ref{sec:results} is devoted to the results. Section \ref{sec:dataanalisis} analyzes the temporal evolution of accident incidence in the city of Barcelona from different perspectives. Section \ref{sec:hazardEstimation}, describes how we predict the impact of urban change on accident rates. Section \ref{sec:netsRisk} describes the way we combine the hazard index estimation with network descriptors to assess the effective risk of pedestrians and vehicles navigating the city. Finally, in Section \ref{sec:conclusions}, we provide the discussion and conclusions of our work.

\section{Data and methods} \label{sec:methods}

In the following subsections we describe the different datasets and machine learning tools that we use throughout the rest of the paper. 

\subsection{Pedestrian and vehicle accident dataset}
\label{sec:accidentDataset}

To study the temporal evolution of accident rates, historical accident records for the city of Barcelona during the period from 2010 to 2017 were downloaded from the city government's open data portal \cite{bcn2019acc}. All data points are geolocated with the GPS coordinates recorded by the local police. Aside from location, we can further distinguish between accidents involving both a vehicle and a pedestrian (simply `pedestrian', or $P$, onwards), and vehicle-to-vehicle accidents (simply `vehicle', or $V$, onwards).  In order to capture the temporal change in accident frequency, each of these groups (pedestrian and vehicle) were further divided according to whether the accidents occurred during the range of years 2010 - 2013, or the range 2014 - 2017 (inclusive). The number of accidents by year and by category is shown in below in Table \ref{tab:datasets_acc}. Note that the total number of accidents increased by about 9.5\% between the two periods. In detail, the total number of pedestrian and vehicle accidents increased by 4.3\% and 10.2\% respectively. Even from this most basic data analysis, it is clear that these traffic safety statistics are moving in the wrong direction. We will provide an extensive study of this dataset in Sec. \ref{sec:dataanalisis}.

\begin{table}[h!]
	\centering
	\begin{tabular}{l|c|c|c|c}
		\hline
		{} & {\bf 2010-2013} &{\bf 2014-2017} & {\bf Entire period }\\
		\hline
		\hline
		Pedestrian Accidents ($P$) & 4478 &  4672 (+4.3\%)    &  9150 \\
		Vehicle Accidents ($V$) & 32031   & 35328 (+10.2\%)  & 67359 \\
		Total & 36509   & 40000   & 76509 \\
		\hline
	\end{tabular}
	\caption{ Accident dataset  disaggregated by accident type and the period in which the accident occurred.}
	\label{tab:datasets_acc}
\end{table}

\subsection{Street view dataset}

\begin{figure}[h]
	\centering
	\includegraphics[width=.95\textwidth]{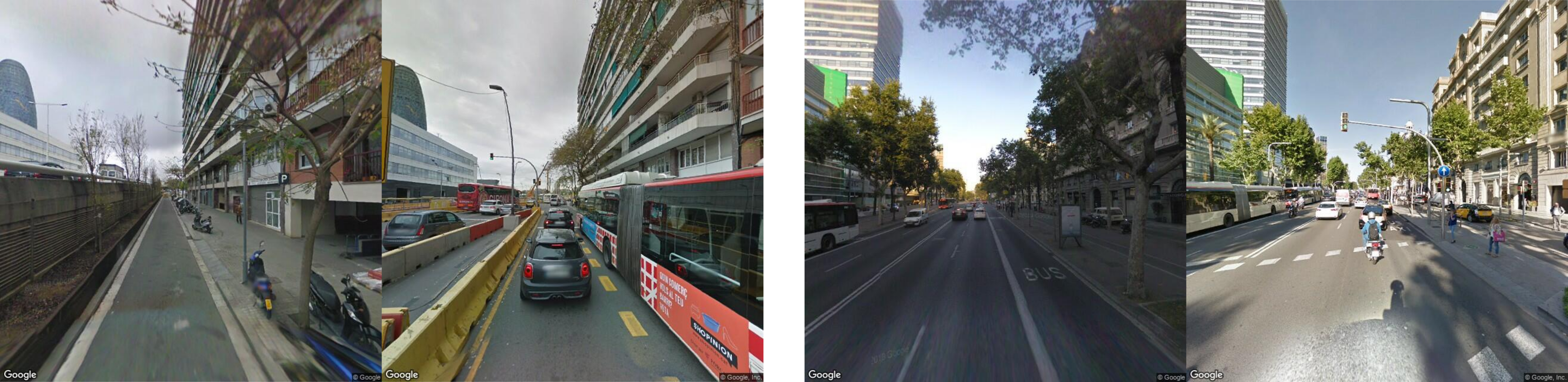} 
	\caption{Illustrative examples of urban interventions that have affected accident rates. Panel A shows a location where the number of vehicle accidents rose from 4 (period 2010-2013) to 10 (period 2014-2017), as new temporary signage and a denser traffic pattern were introduced. Panel B shows a location where the number of pedestrian accidents fell from 9 (period 2010-2013) to 2 (period 2014-2017), in conjunction with the placement of a new crosswalk and a walkable median zone.}
	\label{fig:urbanChange} 
\end{figure}

To construct our Street view dataset, a total of 177K of unique locations from Barcelona's street network, obtained from OpenStreetMap \cite{OpenStreetMap}, were queried from Google StreetView (GSV) \cite{anguelov2010google} to obtain imagery of two years: 2010 and 2017. The querying process resulted in 138K images corresponding for 2010 and 177K images for 2017, resulting in a total of 316K unique images and 138K images pairs from the same location in different time periods, we needed to discard several images from 2017 since no street view image exist in the 2010. These years were selected to provide sufficient temporal distance to capture a large range of urban interventions and changes. Although an exact match between camera viewpoints is nearly impossible, a selection of image pairs across time were checked visually and found to correspond very closely. Image locations along a given street segment are, on average, separated from each other by about 15 meters. GSV images are available for the full 360 degree panorama surrounding the camera. According to the objective of our work, we limited our queries to images facing directly down the direction of traffic, considering two-way streets when necessary. The result of this process is a homogeneous dataset of StreetView image pairs from both periods of study, at every location across the city (Figure \ref{fig:urbanChange} provides some examples). 

The collected images contain GPS locations in their metadata, which, in combination with the historical accident data described in \ref{sec:accidentDataset}, allows us to assign each street image a binary accident category (``safe'' vs. ``dangerous'') according to its proximity to the geo-located accident records. We categorize a point as ``dangerous'' if one or more accidents have occurred with a 50 meter radius of its location. Otherwise, the point is categorized as ``safe''. Table \ref{tab:datasets_img} provides a detailed description of the dataset. It is important to put special emphasis to the high resolution of our tagged image dataset, which can provide information about accident rates and hazard every 15 meters along all the city's road network. 

\begin{table}[h!]
	\centering
	\begin{tabular}{l|c|c|c|c|c}
	\hline
	{} & {} & \multicolumn{2}{c|}{{\bf 2010-2013}} & \multicolumn{2}{c}{{\bf 2014-2017} }\\
	\hline
	\hline
	Barcelona Accidents & Total & Accident  & No accident & Accident  & No accident \\
	\hline
	\hline
	Pedestrian Accidents ($P$) & 316329 & 50945 & 87739  & 62108  &  115537 \\
	Vehicle Accidents ($V$) & 316329  & 101572 & 37112  &  134546 & 43099 \\
	\hline
	\end{tabular}
	\caption{Dataset resulting from the combination of urban images and historical accident data. The table shows the number of locations where we observed at least one accident (``Accident''), and the number where no accidents were observed (``No Accident'). Our data distinguishes between vehicle-vehicle accidents (``Vehicle accidents'') and vehicle-pedestrian accidents (``Pedestrian accidents'').}
	\label{tab:datasets_img}
\end{table}

\subsection{Hazard Index Estimation}

Deep Learning tools for image classification allowed us to develop a hazard index for pedestrians and vehicles (``Hazard Index'') for each StreetView image in the dataset \cite{bustos2021explainable}, adapting a well-known convolutional neural network called Residual Neural Network (ResNet-50-v2) \cite{resnet}. The adaptation includes the removal of the connections from the last layer of the ResNet model, and their replacement with a new, fully-connected layer (with Softmax activation function) and two outputs, according to our number of classes. 

Once the ResNet is trained and ready to draw predictions regarding both accident-related categories ('dangerous' or 'safe'), we define our \textit{Hazard Index (H)} as the probability that an input image is classified as 'dangerous' by the ResNet. Thus, for each street-level input image, the classifiers deliver a value $H$ in the range of [0,1]. This value corresponds to the output of the Softmax activation function of the last fully connected layer of the ResNet architecture. When $H \approx 1$, the location corresponding to the image is considered to be dangerous. On the contrary, when $H \approx 0$, the corresponding location is considered to be safe. Accordingly with the defined accident types ($V$ and $P$), we train our ResNet to estimate two subtypes of Hazard Index: $H^v$ and $H^p$, corresponding to the Hazard Indices for vehicle-to-vehicle and vehicle-to-pedestrian accidents, respectively. Therefore, we end up training 2 models in total over the full datasets, which includes the entire period 2010-2017.

\subsection{Modeling pedestrian and road traffic}
\label{trafficmodel}

It is important highlighting that the described Hazard Index does not explicitly consider the pedestrian and cars volumes navigating the network: Section \ref{sec:netsRisk} addresses these issues. For sake of consistency, in the following we describe the fundamentals we use to estimate vehicle and pedestrian traffic. For a detailed description of the methods and available source code we refer the reader to \cite{rhoads2021sustainable}.

To model pedestrian traffic on top of the sidewalk network of Barcelona, we adapted the method designed by Yang et al.~\cite{yang2014limits}, which assumes a gravitational analogy wherein those locations of the network containing the most services (i.e. having the most mass) are the most attractive to pedestrians. Upon this assumption we can estimate pedestrian Origin-Destination (OD) matrices considering: geographic population data obtained from government Open Data portals, point-of-interest (POI) data from Foursquare~\cite{yang2016participatory,yang2015nationtelescope} (a location-based social network), and real pedestrian flow samples provided by TC Group Solutions (https://www.tc-street.com/). After some numerical calculations, each entry, $o_{ij}$, of the resulting OD matrix (of size $N \times N$ , where $N$ is the number of nodes in the city's sidewalk network) represents the amount of people, per time step, who begin a journey departing from location $i$ to destination $j$. As in
~\cite{sole2018decongestion,henry2019spatio,altshuler2011augmented}, we subsequently fed with our OD matrix to the calculation of the edge betweenness centrality upon Barcelona's sidewalk network to obtain the expected amount of pedestrians that traverse each network edge per time step. That is, the {\it effective} edge betweenness of the pedestrian network, $E_{ij}^{P}$.

The estimation of the road traffic is obtained following a similar procedure, in which the OD matrix is assumed to be all-to-all and the {\it effective} edge betweenness of the road network network, $E_{ij}^{V}$, is obtained considering  Barcelona's road network, weighted by traversal time (edge distance divided by allowed road speed).

\section{Results} \label{sec:results}
\subsection{Temporal evolution of pedestrian and vehicle safety}
\label{sec:dataanalisis}
  
\begin{figure}[h!]
    	\begin{tabular}{ll}
                {\bf (A)} & {\bf (B)}
                \\
                \includegraphics[width=0.48\textwidth]{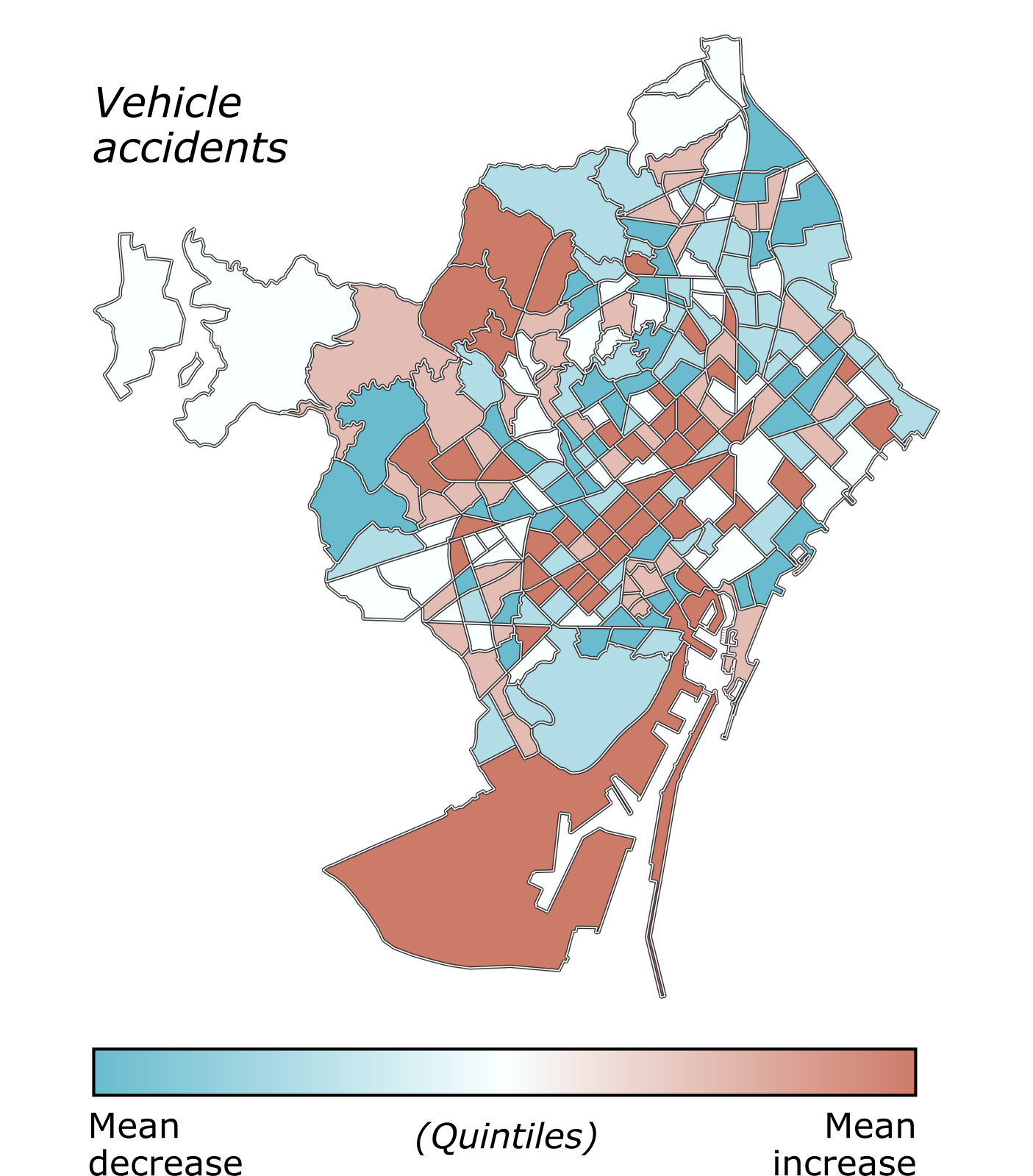}
                &
                \includegraphics[width=0.48\textwidth]{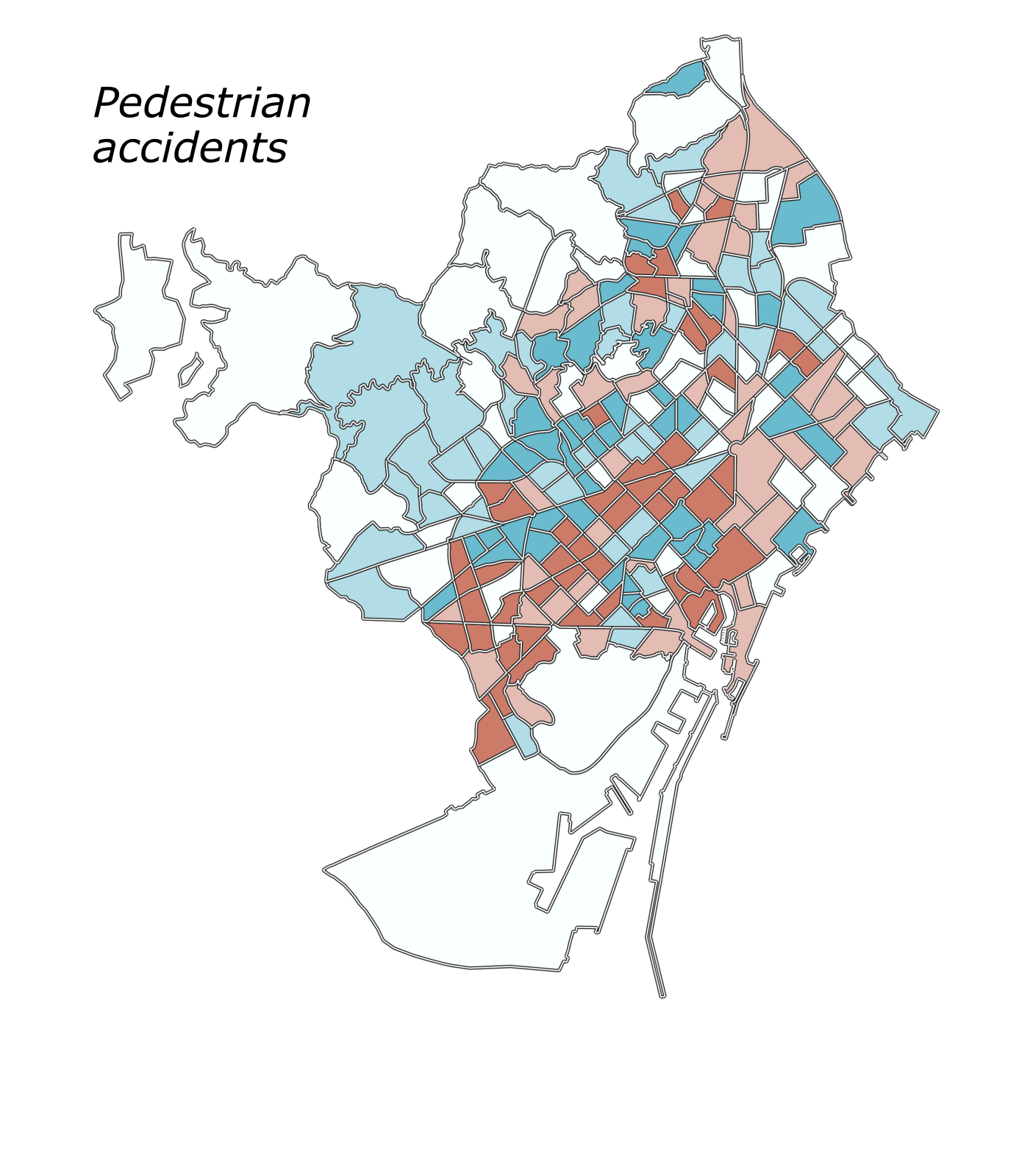}
        \end{tabular}
        \caption{Mean change in number of accidents by neighborhood, disaggregated into vehicle (left) and pedestrian (right) accidents. Besides a general trend of increased accident incidence in the center of the city, the patterns for the two types of accidents seem to show little correlation, either positive or negative.}
        \label{fig:locationAndCorrelationVehiclePedestrianAccidents}
\end{figure}
  
Barcelona's commitment to Vision Zero was formalized in 2017 \cite{bcnVisioCero}. Since then, the city government has implemented several strategies to reduce traffic accidents, including the definition and implementation of a Superblock system \cite{mueller2020changing}, and the use of the so-called tactical urbanism \cite{silva2016tactical}. These initiatives have led to frequent and steady changes in the visual and structural makeup of urban scapes. Some of these interventions have been conducted modifying the urban landscape locally, in order to provide separate areas for vehicle traffic and pedestrian flows. In general, less interaction between pedestrians and vehicles should imply less probability of accident encounters. But, at the same time, increasing levels of population density and traffic flows contribute to the continued conflicting relationship between pedestrians and motorized mobility. 

\begin{figure}[h!]
    	\begin{tabular}{ll}
                {\bf (A)} & {\bf (B)}\\
                \includegraphics[width=0.45\textwidth]{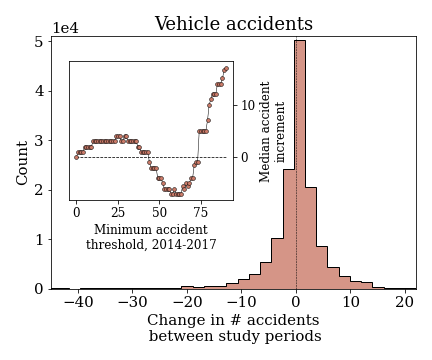}
                &
                \includegraphics[width=0.45\textwidth]{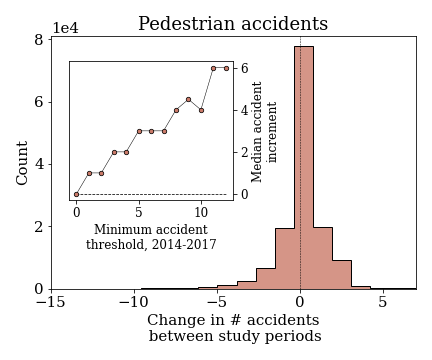} \\
                {\bf (C)} & {\bf (D)}\\                
                \includegraphics[width=0.45\textwidth]{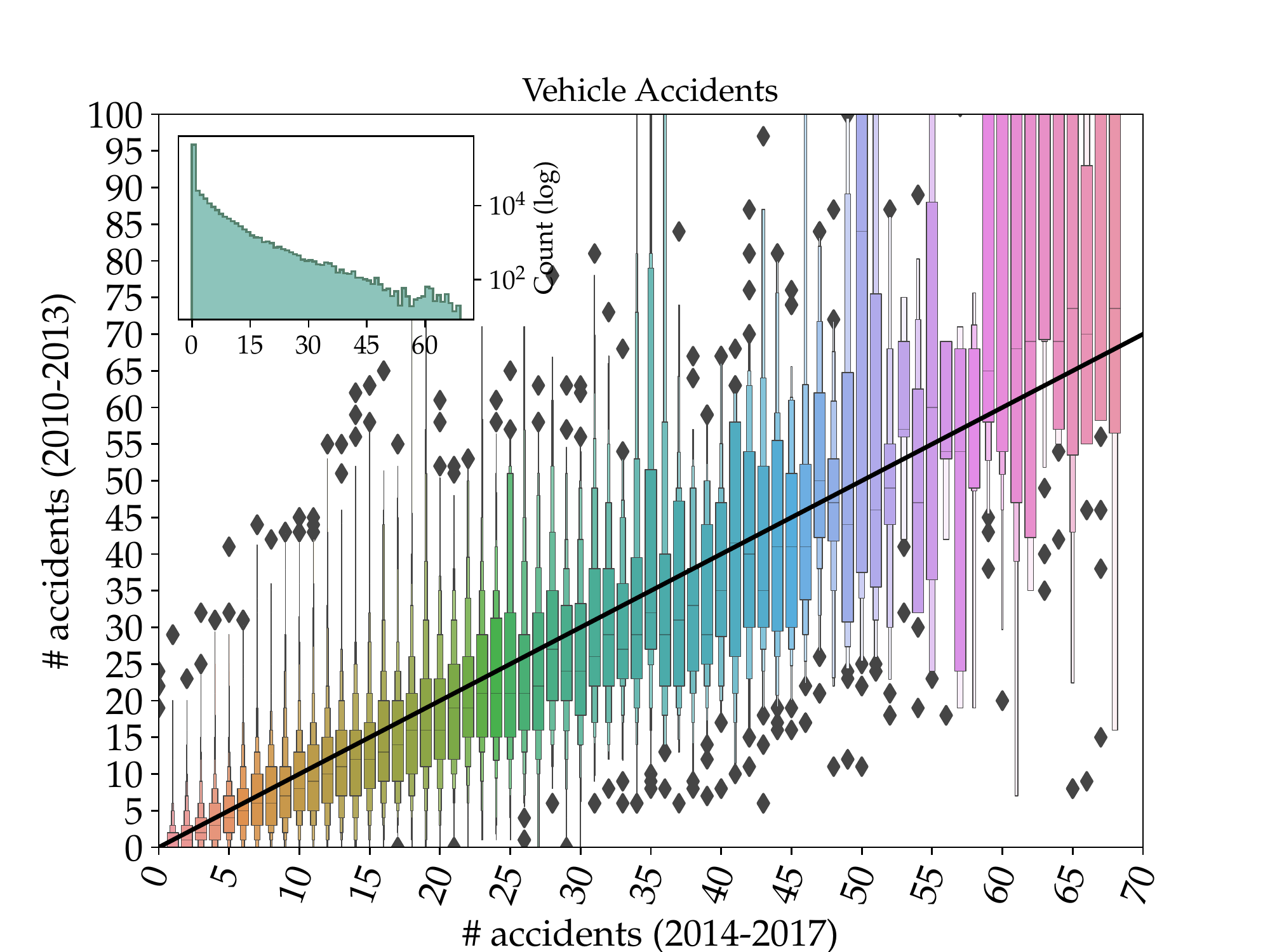}
                &
                \includegraphics[width=0.45\textwidth]{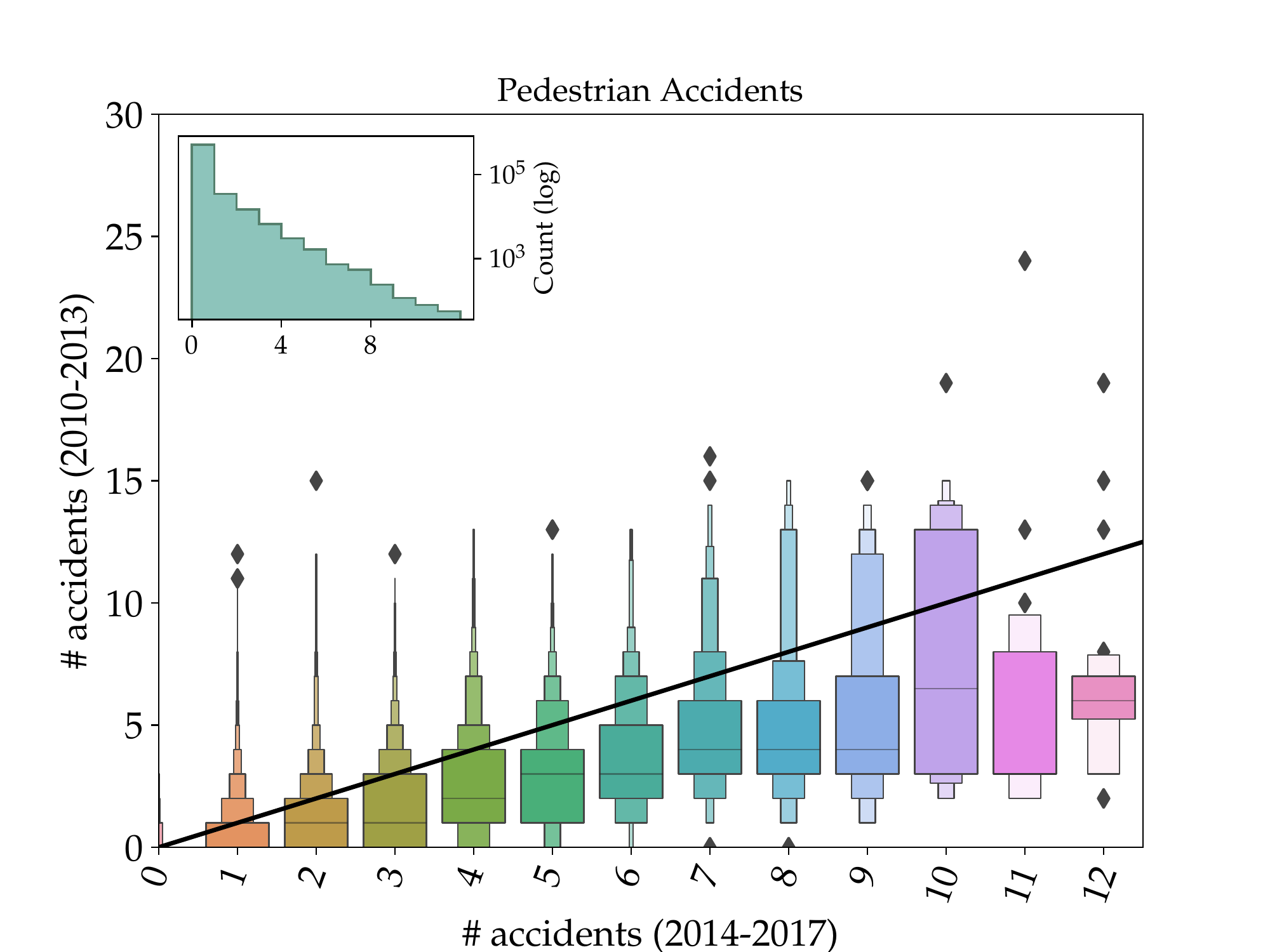}
        \end{tabular}
        \caption{Panels A and B show the variation in the number of accidents, between the two periods of study (2010-2013 and 2014-2017), for vehicle-vehicle and vehicle-pedestrian at each of the 138K image locations of our dataset. Plots display the absolute variation. Positive values indicate an increase in the number of observed accidents. Inset plots show the change (in terms of distribution median) of the number of accidents for points that had $\geq x$ accidents in the period 2014-2017. Panels C and D analyze these variations disaggregated by accident counts on the period 2014-2017 ($x$ axis). Each corresponding box plot represents the distribution of accident counts from the earlier period (2010-2013, $y$ axis). Distribution median and quantiles are represented. Accordingly, when the mean falls below the 1:1 diagonal, these points saw an average increase in accident counts across the two periods. Otherwise, they saw an average decrease. Insets of panels C and D, show the number of points in the dataset with a given number of accidents ($x$-value) in the 2014-2017 period. }
        \label{fig:totalAccidentChange}
\end{figure}

In order to untangle this complicated web of temporal cause and effect, we first provide a descriptive analysis of the evolution of urban accidents involving pedestrians, vehicles, and a combination of both. Figure~\ref{fig:locationAndCorrelationVehiclePedestrianAccidents} shows that the evolution in the number of accidents does not seem to follow any particular spatial pattern. Increases and reductions are distributed almost homogeneously across the city. Additionally, we observe that there is very limited correlation between where vehicle-vehicle and vehicle-pedestrian accidents occur, most likely reflecting the differences between pedestrian and vehicle mobility dynamics. Panels A and B of Fig.~\ref{fig:totalAccidentChange} and Table~\ref{tab:datasets_acc} evidence quantitatively that accident incidence for both pedestrians and vehicles has increased over the course of the two periods in absolute number, but also, in some districts, on a point-by-point basis. Aside from this slight average increase, we also observe that the distributions of the variation in accident incidence is quite broad, see panels A and B of Fig.~\ref{fig:totalAccidentChange}. Some locations have experienced a major improvement, while others a major deterioration. In panels C and D of Fig.~\ref{fig:totalAccidentChange} we examine the evolution in accident incidence disaggregated by the number of accidents in the period 2014 to 2017. It is interesting to note that, on average, pedestrian safety has deteriorated independently of the present situation. In contrast, for vehicle-vehicle accidents, although the trend is similar for areas with fewer accidents, heavily affected points (more than 50 accidents observed) have, on average, seen an improvement in safety. Inset on panels A and B of Fig.~\ref{fig:totalAccidentChange} provide statistics on the variation of accident incidence when considering locations with a number of accidents greater than a given threshold (x-axis value). Over all, the average behavior indicates that interventions from local authorities seem to be insufficient to contain road safety, and the large deviation observed also indicates that the problem seems to require a multi perspective analysis to prioritize interventions.  

The preceding analysis provides a global picture of the situation, but more work is needed to uncover the connections in the evolution of accident incidence over time at particular locations, and how this relates to interventions and changes in the built environment.

\subsection{Predicting the impact of urban interventions on pedestrian and vehicle safety}
\label{sec:hazardEstimation}

The advantage of using automated tools to aid in the design of safe urban environments comes from those tools being able to inform whether a location will increase or decrease accident incidence as a result of a urban intervention or change in its physical layout. 
Currently, this process is usually done by relying on specific field knowledge (e.g. urban planers, municipalities, etc.), or through {\it a posteriori} statistical analyses of the effects of interventions. However, both methods have their own downsides. On the one hand, specific knowledge requires time and experience, which is harder to obtain as cities grow; on the other, statistical analyses require accidents to occur, which, besides being undesirable, does not allow to assess the safety of those locations where accidents do not occur that often. 

In this section we will locally address this issue in an automated way. Specifically, we assess whether the difference between the Hazard Index estimated from Street View images across both periods of study for the same location is a good proxy to predict the empirical change in accident incidence at that location. Recall that the trained neural network is only fed with Street View images: no information about traffic, pedestrian density or time is provided. Thus, differences in the Hazard prediction are obtained solely considering the physical or structural change in the scene.

\begin{table}[h!]
	\centering
	\begin{tabular}{|l|c|c|c|c|c|c|c}
	\hline	
	Type & \specialcell{Incr.\\Acc.}  & \specialcell{Incr. Acc.\\ \&  \\Incr. Hzrd} &\specialcell{Incr acc \\ \& \\ Incr. Hzrd $\pm 0.05$}  &  \specialcell{Decr. \\Acci}  & \specialcell{Decr. Acc\\ \& \\ Decr. Hzrd}   &  \specialcell{Decr. acc \\ \&  \\ Decr. Hzrd $\pm 0.05$}\\
	\hline
	\multicolumn{7}{l}{Full dataset}\\	
	\hline
 	$P$ & 30230 & 18533 (61\%) & 21842 (72\%) & 30672 &  18557 (60\%) &  22125 (72\%)  \\
 	$V$ & 58230  & 36372 (62\%) & 45673 (78\%) & 48486 &  23085 (48\%) & 32926 (68\%) \\	
	\hline
	\multicolumn{7}{l}{Restricted dataset}\\
	\hline
 	$P$ & 12630 & 9060 (71 \%) & 10661 (84\%) & 12975 & 9156  (70 \%) & 10884 (84 \%)  \\
 	$V$  & 36827 & 23313 (63 \%) & 31154 (85\%)  & 31729 & 15536 (49\%) & 24265 (76\%) \\
	\hline	
	\end{tabular}
	\caption{Evaluation of the correlation between the increment/reduction of the number accidents with the increase/decrease of the Hazard Index. For each of the well estimated image-pairs in which the number of accidents has increased (or decreased), we counted those in which the Hazard Index also increase (or decreased). Columns including $\pm 0.05$ provides the same statistics. These results provides us with information about the quality in detecting large variations on accident incidents, or else in detecting no variations. Results labelled as {\it Full dataset}, regard the analysis of the 138k image pairs we could obtain for Barcelona. Results labelled as {\it Restricted dataset}, regards results obtained considering the subset of images correctly classified by the method in \cite{bustos2021explainable}.}
	\label{tab:accuracy_correlation_hazard_acc}
\end{table}

Results in Table \ref{tab:accuracy_correlation_hazard_acc} show the performance we obtain in predicting the increase or decrease of accident rates. Results considering all data available show that we can predict an increase or decrease on accident rates with an accuracy of 60\%, which increase to 70\% if we consider a small tolerance on the prediction. Since we know {\it a priori} that error rates in predicting dangerous vs. not dangerous images should be approximately 24\% per image \cite{bustos2021explainable}. We could estimate that in 42\% of image pair either at least one of the images will be wrongly classified. To understand how this affects our predictions we also evaluated an scenario in which only the set of image pairs where both images could be classified correctly in a pre-processing step using the method in \cite{bustos2021explainable} are used. Accuracies in this scenario raise to 70\% and 80\%, which is promising. 

Overall results in Table \ref{tab:accuracy_correlation_hazard_acc} show that method works good in practically all scenarios analyzed, except in predicting the decrease in accident rates for cars.

\begin{figure}[h!]
    	\begin{tabular}{ll}
                {\bf (A)} & {\bf (B)}
                \\                
                \includegraphics[width=0.45\textwidth]{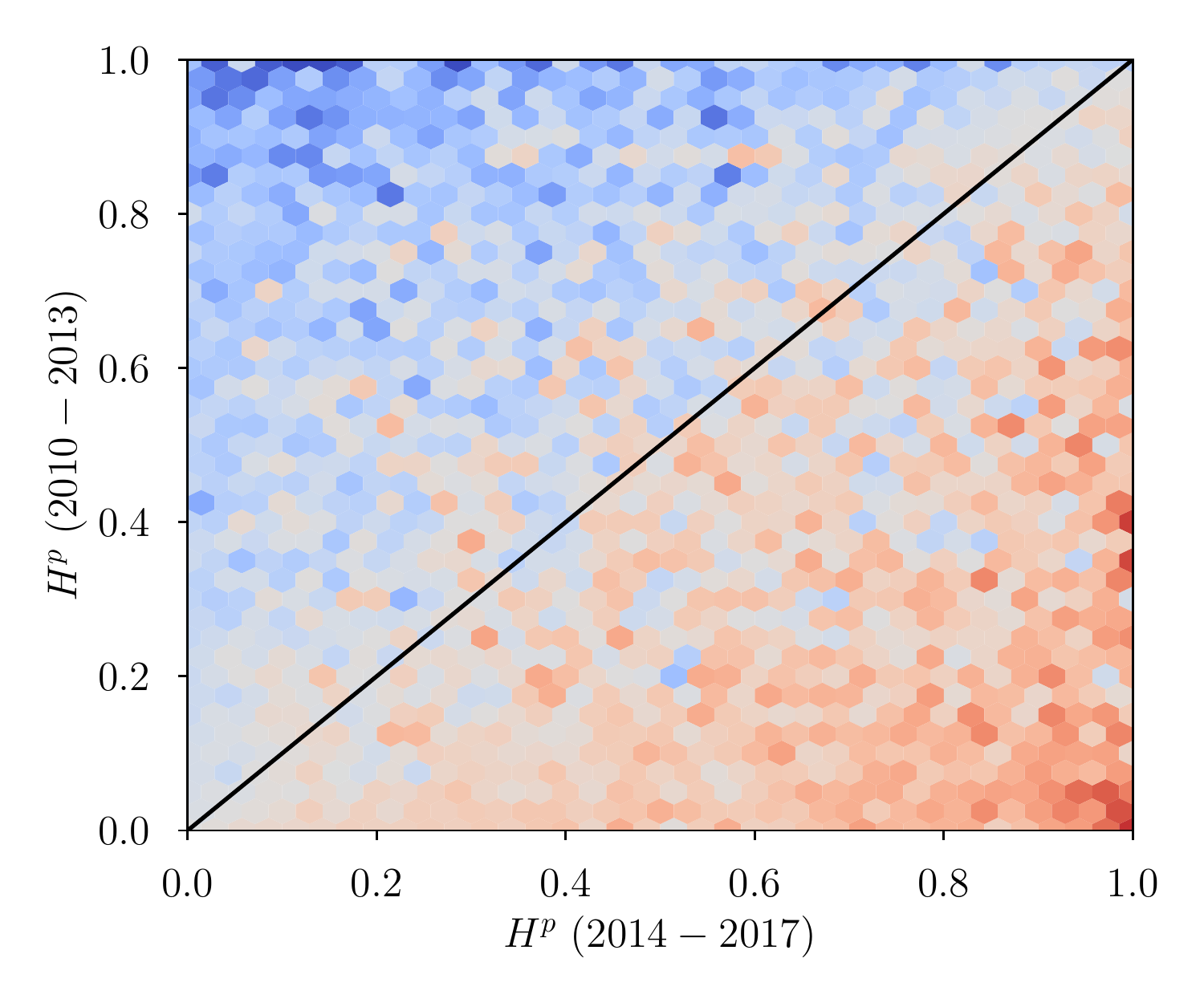}
                &
                \includegraphics[width=0.45\textwidth]{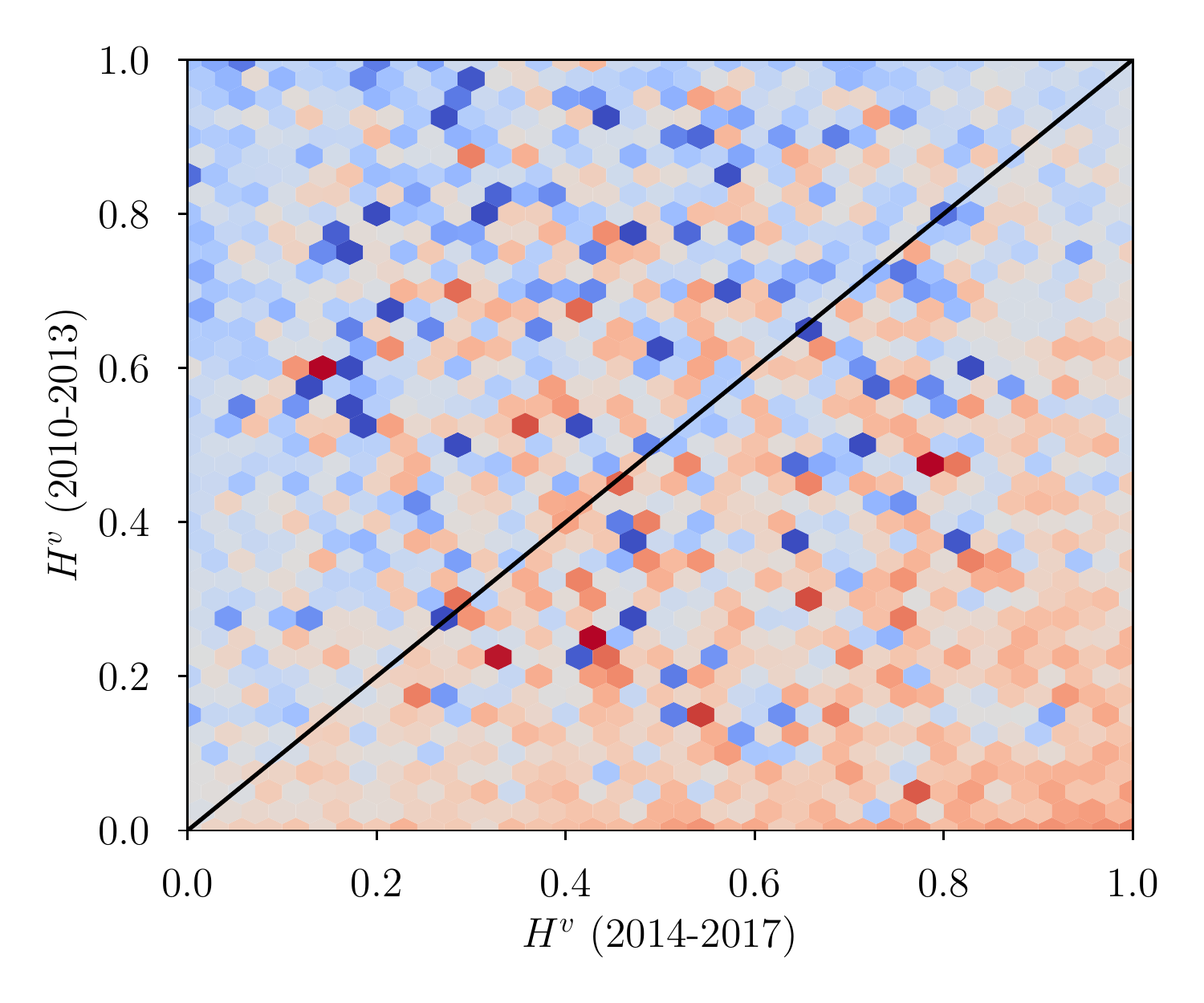}
        \end{tabular}
        \caption{Comparative analysis of the observed change in the number of accidents and the change predicted by the Hazard Index for the same location across the two periods of study. The plot has been assembled considering, for each location, the number of accidents and the Hazard Index prediction from the first (2010-2013) and second (2014-2017) periods. The color of each hexagonal bin represents the change in number of accidents for all locations with particular Hazard Index pairs, $(x,y)$ coordinates representing the different periods of study. Red tones indicate that accidents increased, and blue tones accidents decreased. Panel A shows the results for vehicle-pedestrian accidents, and panel B shows those for vehicle-vehicle accidents.}
        \label{fig:h_real_accidents}
\end{figure}

\begin{figure}[h!]
    	\begin{tabular}{l}
		{\bf (A)} 
		\\
		\includegraphics[width=0.8\textwidth]{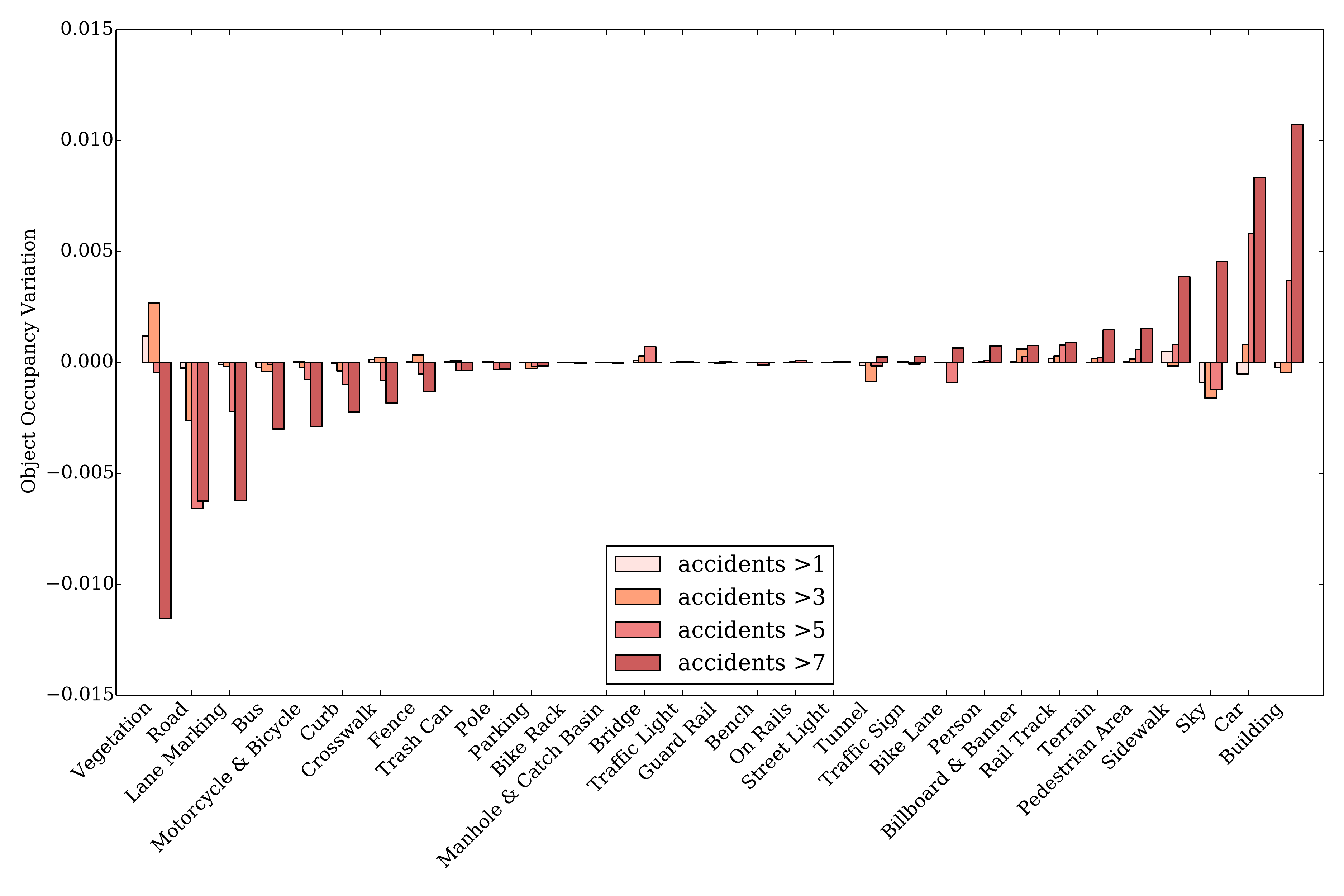}
		\\
		{\bf (B)}
		\\
		\includegraphics[width=0.8\textwidth]{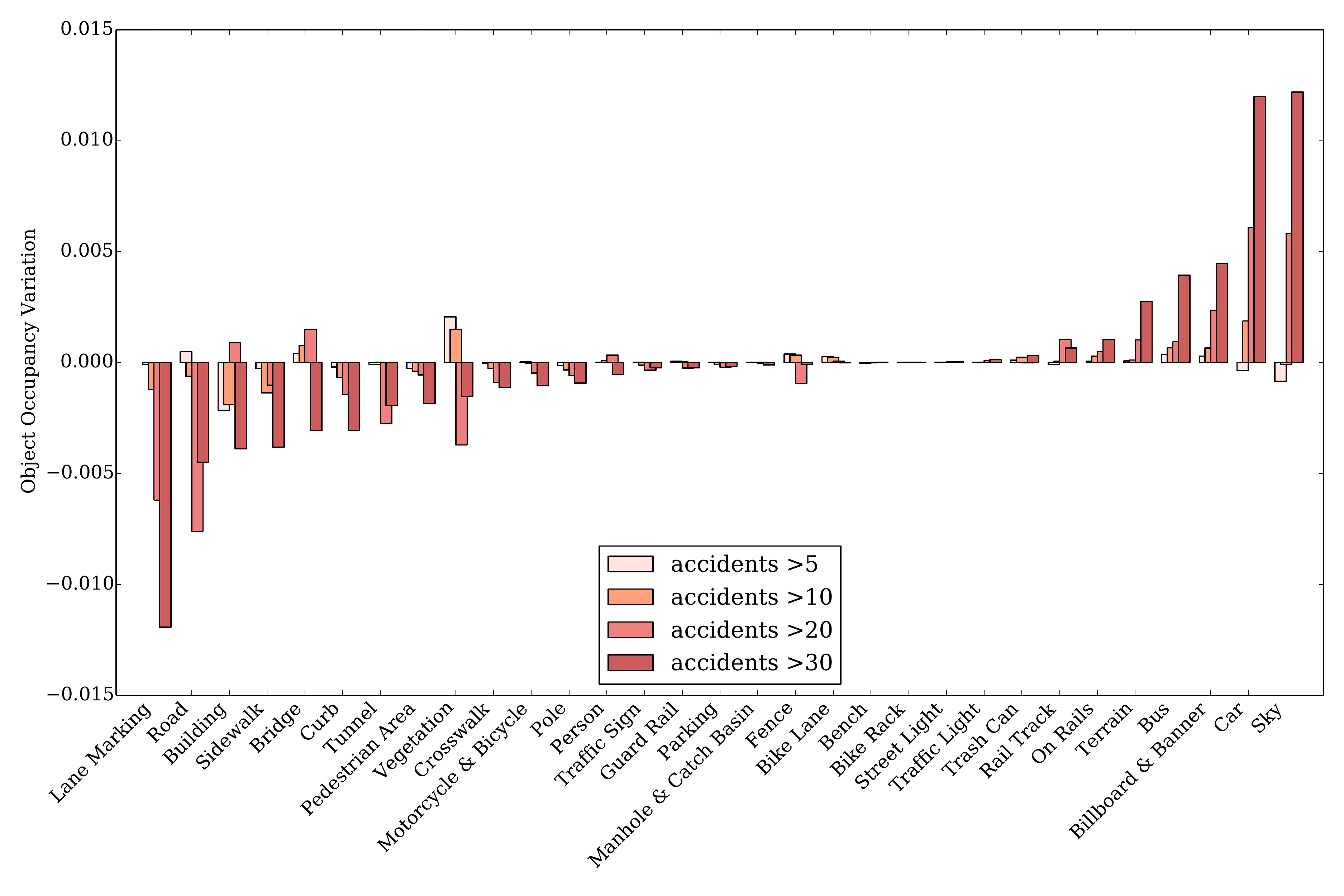}
        \end{tabular}
        \caption{Comparative analysis of urban object occupancy variation with respect to variation of accident rates. For each pair of images, taken in the same place, we calculated the difference between the number of accidents and then clustered all pairs of image according this difference. Subsequently, we assessed which objects gained or lost importance with respect to accident rate variations. Specifically, being  $v(i) \in \mathbb{R}^{C}$, a vector containing information of the relative fraction of object category $C$ in image $i$. The mean difference is calculated $ (v(i)_{m} - v(i)^{\ell}) / N $ , where $m$ refers to images with more accidents, $\ell$ to less dangerous images, and $N$ the number of pair-images with this condition. Panel A presents the results for pedestrians and Panel B for vehicles. To obtain the object occupancy, we first segmented all images on our dataset using the Inplace-ABN (DeepLabV3+WideResNet-38) implementation \cite{bulo2018place}, already trained with the Mapillary Vistas dataset \cite{neuhold2017mapillary}. Once segmented, object occupancy, in percentage of pixels, was obtained for each object in each image.}
        \label{fig:mean_delta_seg}
\end{figure}

Figure~\ref{fig:h_real_accidents} explores these results with higher detail. We see that, on average, the larger the increase (decrease) in the Hazard Index (lower-right and upper left corners of the plot) the larger the increase (decrease) in accident incidence. This implies that, beyond predicting in a binary fashion whether or not there will be a variation in accident incidence, the Hazard Index is able to provide information about the magnitude of such changes. To complement the previous results, Fig.~\ref{fig:mean_delta_seg} shows that the presence of some objects related to larger accident rates become more predominant in increasingly dangerous locations. This provides more evidence for our hypothesis that the increasing frequency of some urban objects may be directly related to accident incidence.

\subsection{Assessing the risk evolution of sidewalks and road networks}
\label{sec:netsRisk}

The analysis provided in the previous sections does not explicitly include information regarding the demand of the corresponding transportation networks: sidewalks and roads. Although very informative about the intrinsic hazard of the scene, disregarding the network perspective and demand, our localized predictions may be insufficient to quantify the effective hazard population experience on their daily mobility and, consequently, useless for efficient prioritization of urban interventions. Note, for instance, that a location with large hazard index but with practically no traffic (pedestrians or vehicles) may not be too problematic. At the same time, a relatively less dangerous location with numerous trajectories passing through may result in a large accident rate. In this section, we combine the hazard index of sidewalks and road segments with network features to provide a practical analysis of urban safety. 

\begin{figure}[h!]
    	\begin{tabular}{ll}
                {\bf (A)} & {\bf (B)}
                \\                
                \includegraphics[width=0.5\textwidth]{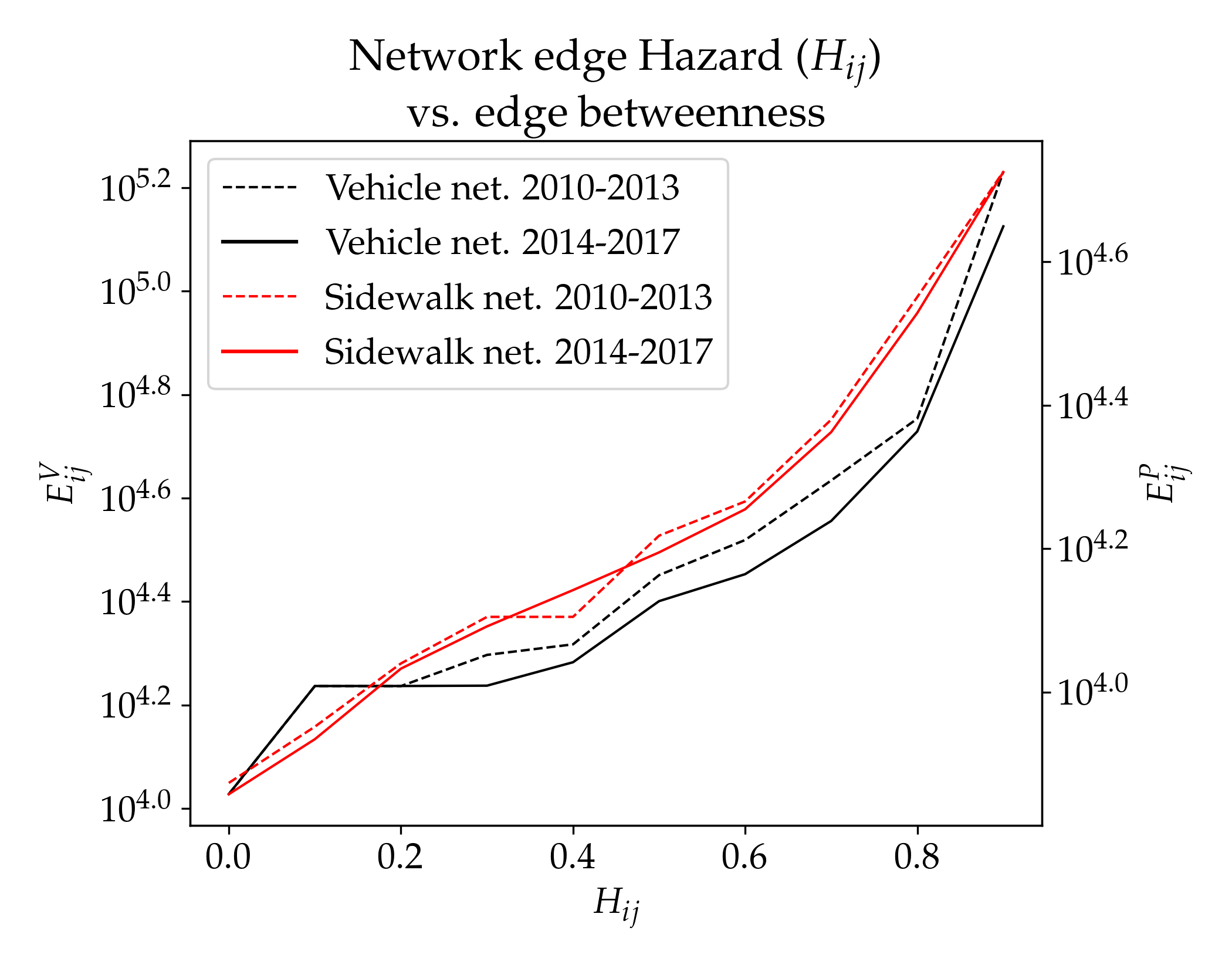}
                \includegraphics[width=0.5\textwidth]{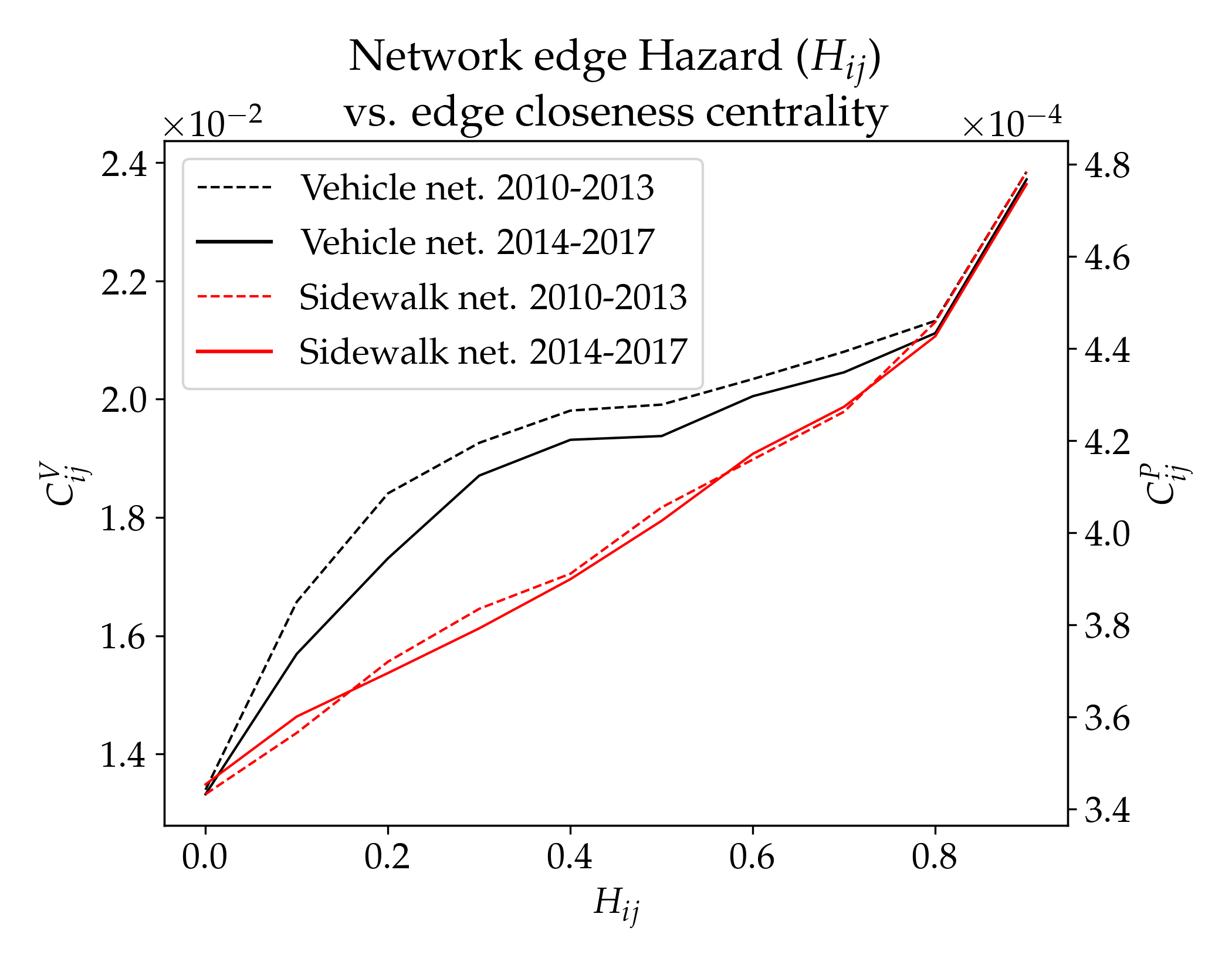}
        \end{tabular}
        \caption{ Correlation between Hazard Index and network centrality measures, across periods and accident type. Panel A shows the results regarding betweenness centrality and Panel B regarding closeness centrality.}
        \label{fig:correlation_hazard_effectivehazard}
\end{figure}

We start by examining the relation between the estimated Hazard Index, and centrality (considering here two flavours: closeness \cite{newman2009networks} and {\it effective} edge betweenness \cite{freeman1977set}) of the corresponding sidewalk and road networks. Betweenness centrality is representative of the traffic demand on a transportation networks when correctly fitted with real data, see Sec.~\ref{trafficmodel} and \cite{rhoads2020planning}. Closeness centrality, a purely structural measure to describe network nodes, is useful to determine the geometric centrality of a network. Since our hazard index is related to the edges of the network, we construct an edge version of closeness centrality, $C_{ij}$, by averaging the closeness of an edge's terminal nodes.

Results in Fig.~\ref{fig:correlation_hazard_effectivehazard} evidence a clear positive relation between betweenness centrality and the Hazard Index, meaning that sidewalks and roads with high demand are also the most dangerous ones. Results regarding closeness centrality are qualitatively equivalent for vehicles and pedestrians. 

\begin{figure}[h!]
    	\begin{tabular}{ll}
	        {\bf (A)} & {\bf (B)}\\
                \includegraphics[width=0.50\textwidth]{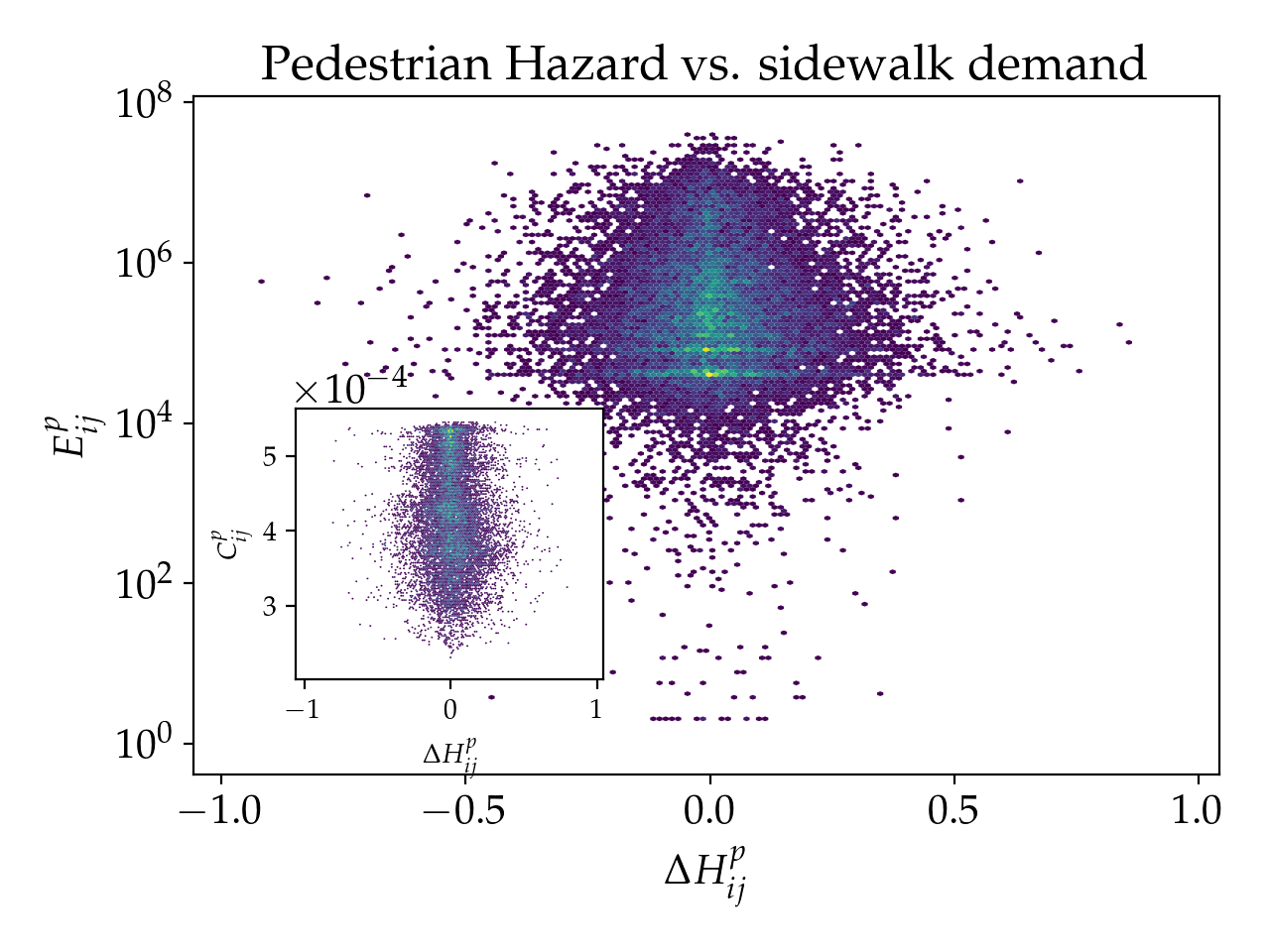}
                &
                \includegraphics[width=0.50\textwidth]{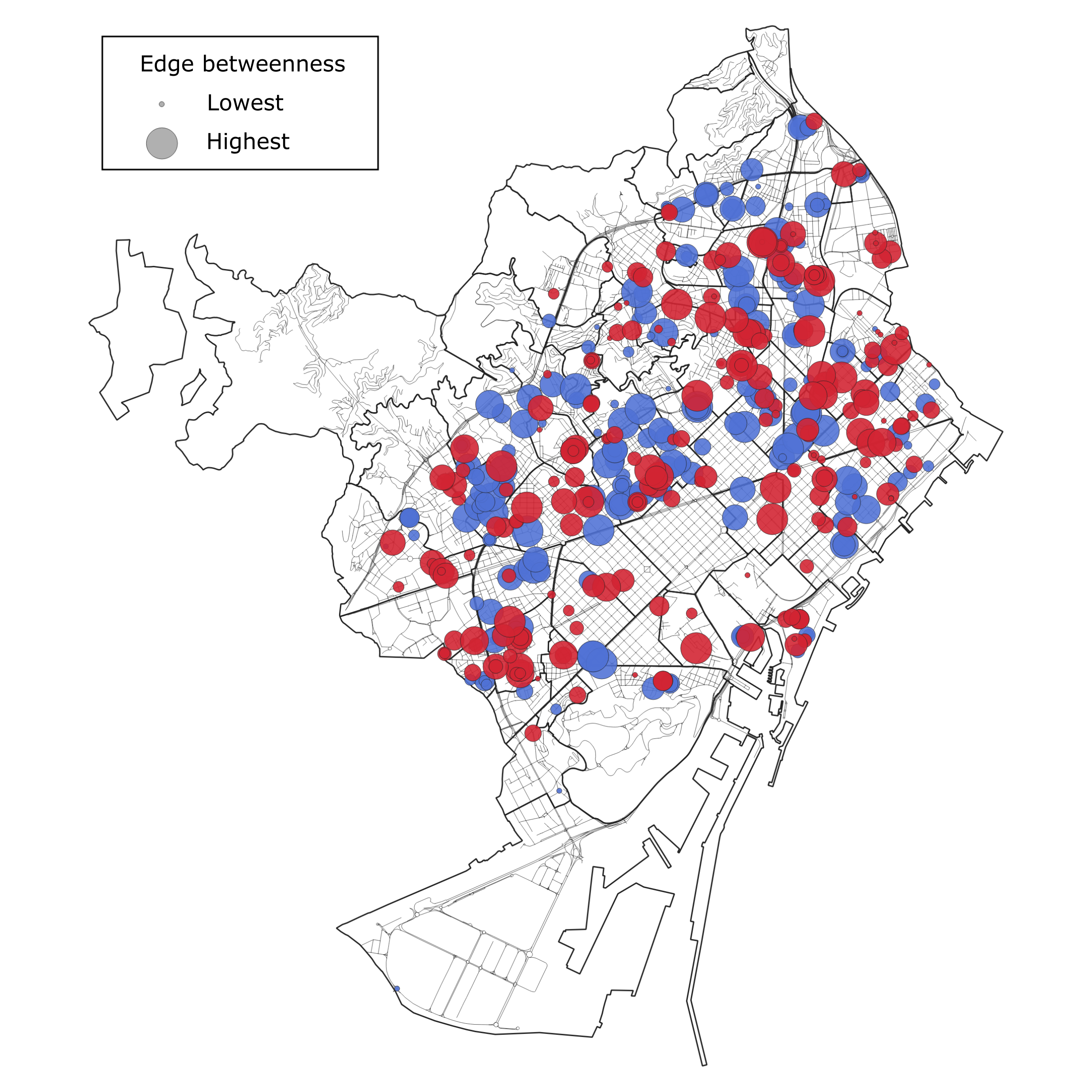}
        \end{tabular}
        \caption{ (A) Correlation plot with the variation of the Hazard Index, across periods, with respect to the betweenness centrality of the same network segment. (B) Location over the map of Barcelona of the sidewalk segments with larger variation (increase or decrease) of Hazard Index. These points correspond to those segments with Hazard Index above or below $2\sigma$ with respect to the average variation.}
        \label{fig:evolutionEffectiveHazardIndex}
\end{figure}

\begin{figure}[h!]
    	\begin{tabular}{ll}
		{\bf (A)} & {\bf (B)}
		\\                
		\includegraphics[width=0.45\textwidth]{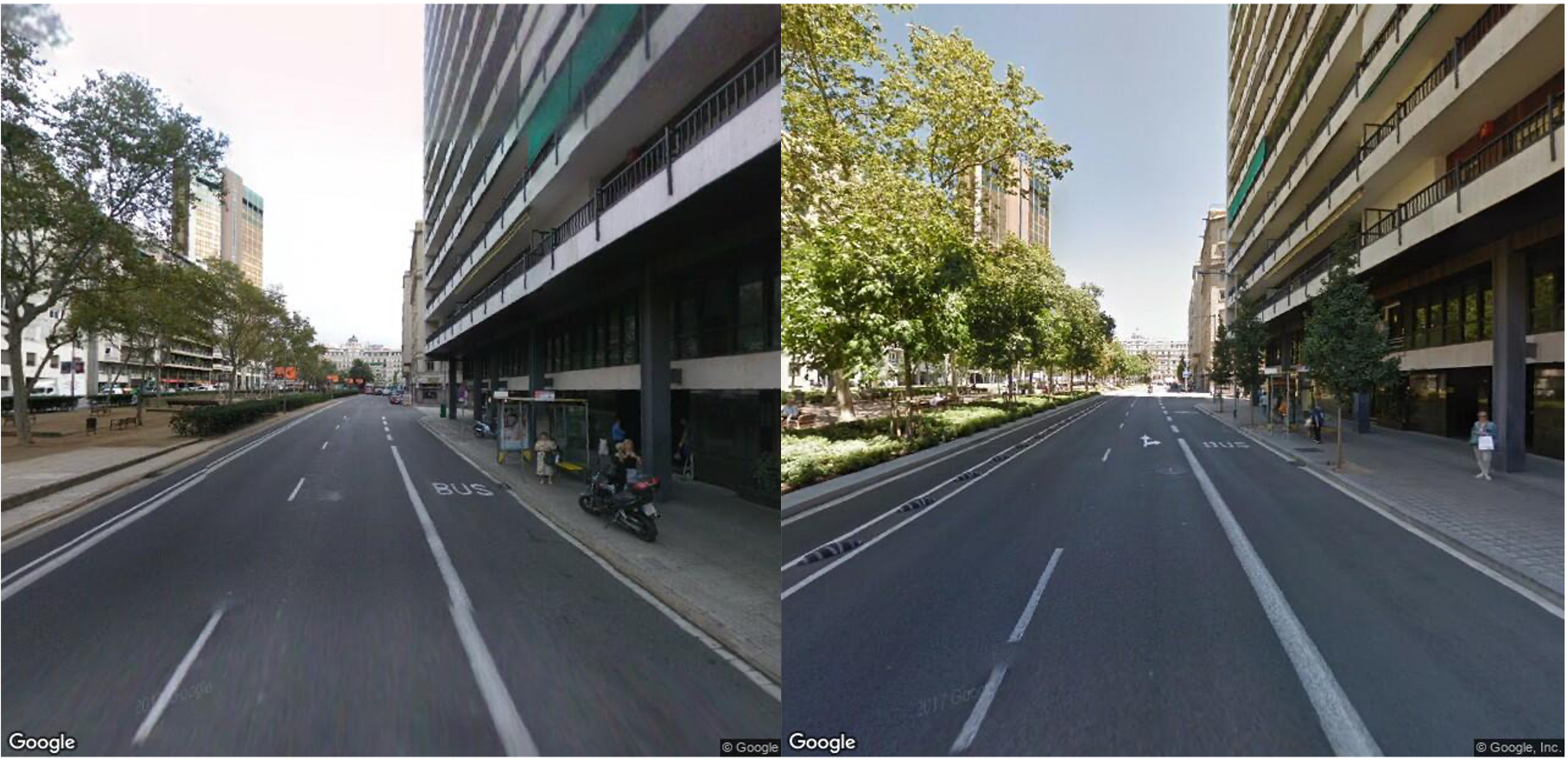}
		&
		\includegraphics[width=0.45\textwidth]{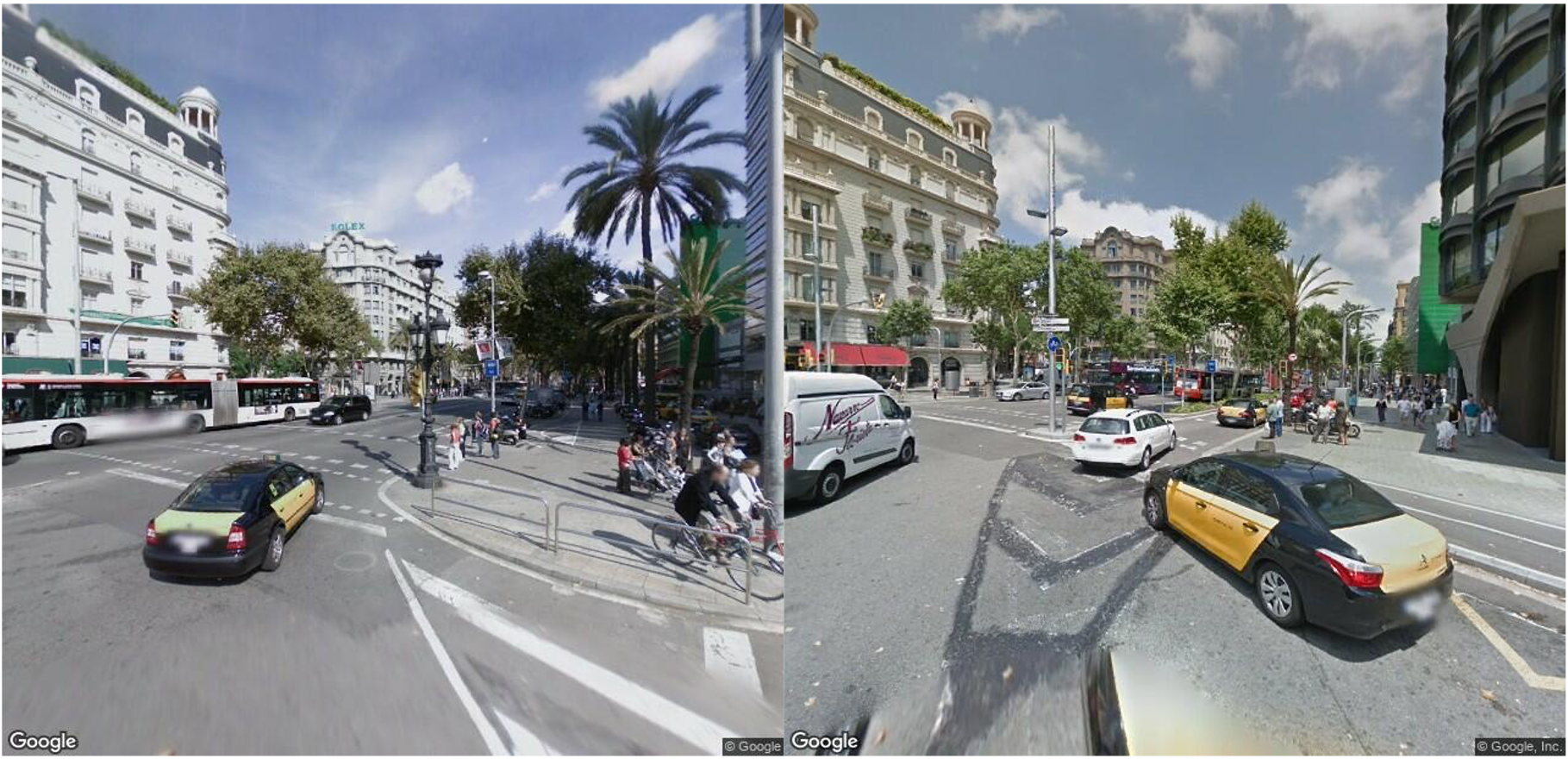}
		\\
		{\bf (C)} & {\bf (D)}
		\\
		\includegraphics[width=0.45\textwidth]{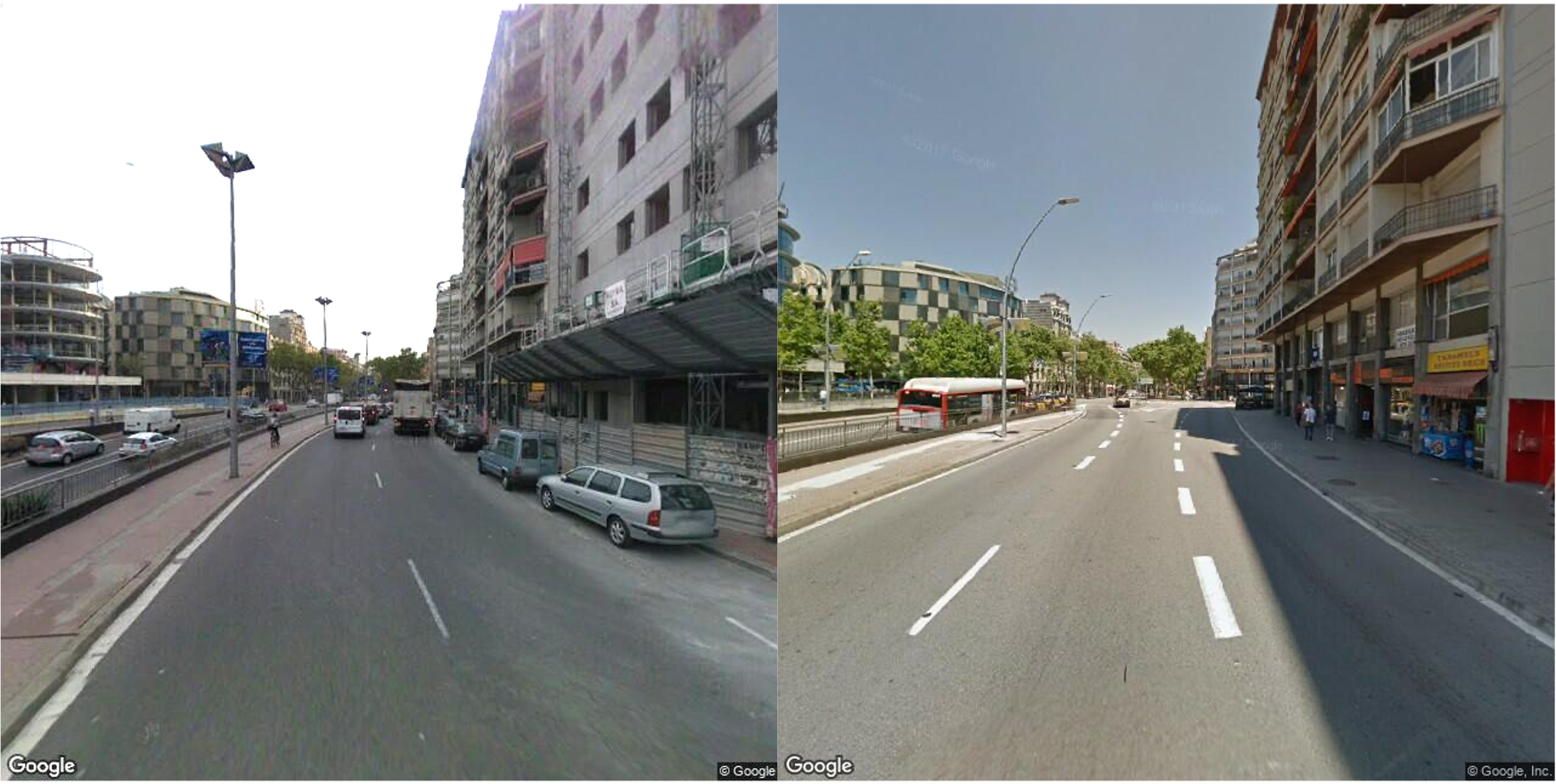}
		&
		\includegraphics[width=0.45\textwidth]{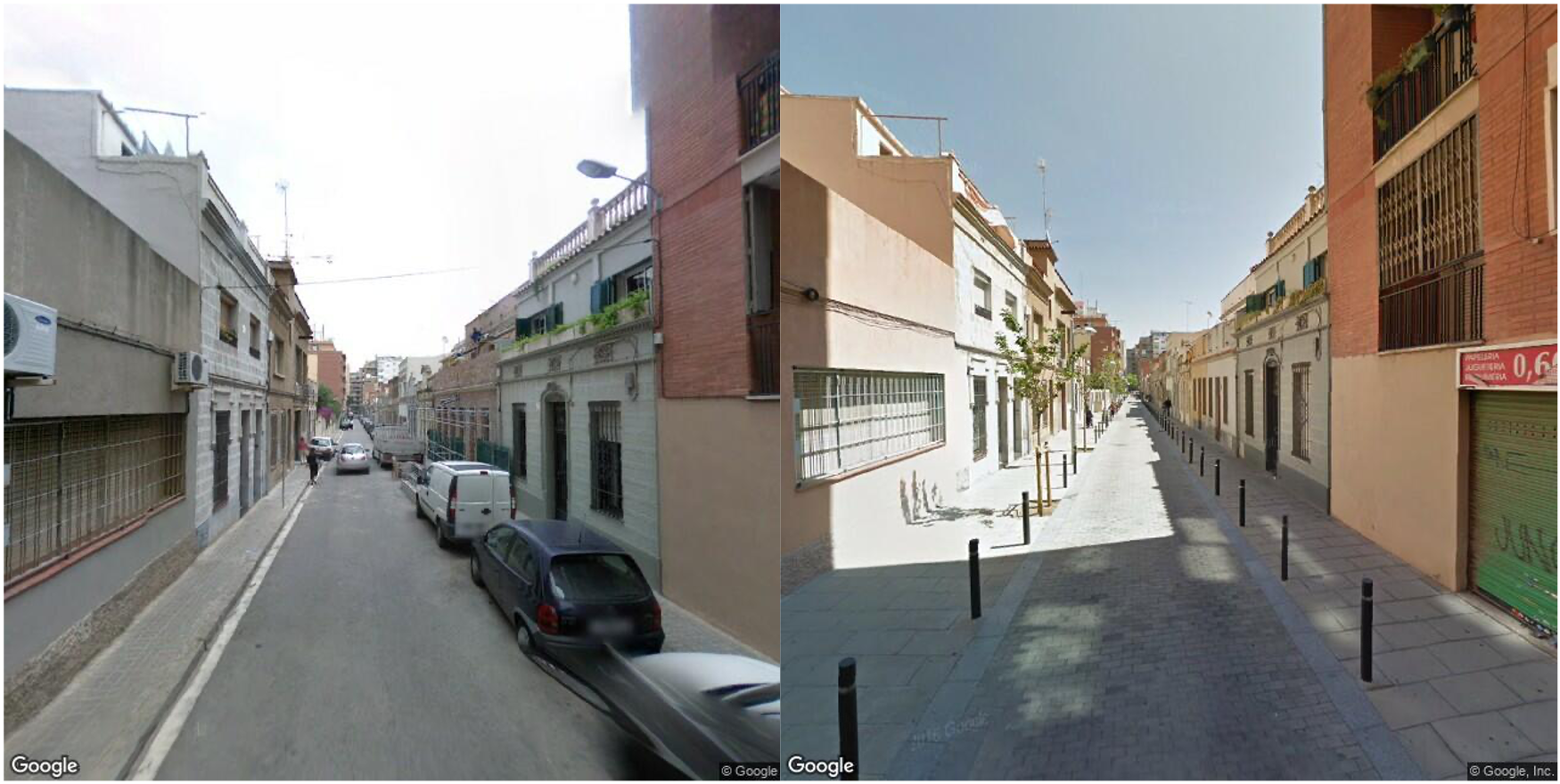}  
        \end{tabular}
    \caption{Examples of urban scenes with different effective betweenness values with large increase or decrease of Hazard Index across periods. For each pair of images, left side one regards the period 2010-2013 and right side one to the period 2014-2017. Images on the top row (panels A and B) regard images obtained from pedestrian areas with large betweenness, which contrast with the images corresponding to low betweenness areas seen in the bottom panels (panels C and D). Left panels (A and C) correspond to images where we observed a decrease in the hazard index and right panels (B and D) a large increase. Panel A represents a location where pedestrian accident rates reduce from 4 to 1 (hazard reduction from x to y). In the original urban setup a park on the left side of the image was accessible by crossing the road from a non-authorized area, which probably was encouraging pedestrians to do so. We see that the urban actuation has modified the area to physically forbid pedestrian crossings, adding, additionally, a bike lane. In panel B, we see a location where pedestrian accident rated have increase from 2 to 10. We can see different interventions that may have triggered such increase. We highlight two. The first regarding the fence protecting pedestrians have been removed and the second regarding the reduction of distances between sidewalks crossing the big road (small pedestrian island has been introduced), which may encourage pedestrians to rapidly cross the road while traffic lights in read. Panel C shows a low pedestrian betweenness that have increased accident rates from 2 to 5, see that several services has been made accessible in the area, augmenting probably the affluence of pedestrians \cite{yoshimura2021street}. Finally, Panel C shows a region with low pedestrian betweeness where accident rates have been reduced from 4 to 0. See that interventions have been addressed to separate sidewalk from the road network.}
	\label{fig:objects_delta}
\end{figure}

We finally assess how the Hazard Index has varied across periods with respect to the network centrality measures. We do not observe a clear prioritization of interventions neither in terms of hazard nor traversal demand, see Fig.~\ref{fig:evolutionEffectiveHazardIndex} panel A. Additionally, Fig.~\ref{fig:evolutionEffectiveHazardIndex} panel B shows the distribution of the locations with larger variation (the largest 5\% increase or decrease) of the Hazard Index contrasted by its betweenness value. Distribution of improvements of the Hazard Index is homogeneously spread over the urban landscape. This seems to indicate as well an homogeneous distribution of interventions across districts. In turn, deterioration of safety also seems to be homogeneously distributed. Fig.~\ref{fig:objects_delta} shows some illustrative examples of locations with different betweenness values that experienced a large variation of Hazard Index (increase or decrease). Finally, it is worth mentioning that transportation demand (betweenness) is poorly related to the registered urban interventions, either decreasing or increasing the Hazard Index.

\label{objects}

\section{Discussion, limitations and conclusions}
\label{sec:conclusions}

As cities grow, their management becomes increasingly difficult. Many actors (urban planners, data scientists, local administration, etc.) need to coordinate to make sense of urban data, to design and prioritize interventions in order to reach a kind of "urban efficiency" understood in a broad manner (efficient, sustainable and equitative mobility, no traffic accidents, economic growth, etc.). At the same time, many tasks may be automated to some degree by means of urban sensors and big data analysis. In line with other recent advances which take profit of new and extensive data-sets \cite{daraei2021data,bogacz2021modelling,yoshimura2021street,naik2014streetscore}, machine learning ~\cite{ibrahim2020understanding, he2021inferring, bustos2021explainable,he2021inferring,xu2020towards} and their combination with complex networks tools \cite{mahfouz2021road,sargoni2020sequential,alhazzani2021urban,rhoads2021sustainable,mukoko2019examining,rifaat2011effect}, our work focuses on the automated assessment and improvement of urban traffic safety, for both pedestrian and vehicles. 

Here, taking these conclusions further, we have shown that our methods are sensitive enough to predict the effect urban interventions on safety, for both pedestrians and vehicles. This finding opens the door to the automated and simulated assessment of the effects of specific interventions in the built environment, potentially allowing planners and stakeholders to gain insights on the effectiveness of proposed actuations.

Clearly, other factors not considered here may have an impact on the evolution of accident rates: changes in vehicle or pedestrian flows, new services in particular areas, or simply different weather conditions. While our deep learning pipeline can be easily modified to include other kinds of data, the results we present in this paper show that visual information alone is enough to achieve high levels of accuracy. This is important, since driver point-of-view images (like those we use) are readily available at low cost, particularly in light of new open providers of imagery, such as Mapillary \cite{mapillary2019}. This means that no costly data acquisition phases are needed to obtain predictions for practically any city world-wide. 

We hope that, encouraged by the good accuracy we have obtained, our work can stimulate the development of more sophisticated approaches to the problem of automatically proposing improvements in the urban built environment, for example with the use of Generative Adversarial Networks to produce virtual, plausible alternatives to target scenes (a kind of automatic ``safe-ification''). These techniques could be further trained to estimate the potential cost of possible interventions, also considering the trade-off between cost and gains in safety, in order to aid urban planners in the prioritization of urban interventions.

Finally, we would also like to stress the importance of our work in contributing to the design of pedestrian friendly and sustainable cities. The design of efficient \cite{de2015personalized}, enjoyable \cite{kauer2018mapping}, and safe cites remains a challenge with international attention, listed in the Sustainable Development Goals of United Nations. Within the development of Goal 11 (Sustainable cities and communities), some of the objectives rely on the definition/characterization and analysis of pedestrian paths, from the amount of existing services (or points-of-interest)\cite{yang2014limits}, to the width of sidewalks \cite{rhoads2020planning}. The automatic characterization of path segments is crucial to the automated design of these routing strategies \cite{xu2020towards,de2015personalized}. Our work, with its ability to draw very detailed (microscopic) safety predictions, contributes to one of those urban aspects that have historically been considered important: pedestrian safety.




\bibliographystyle{bmc-mathphys} 


\begin{thebibliography}{63}
\ifx \bisbn   \undefined \def \bisbn  #1{ISBN #1}\fi
\ifx \binits  \undefined \def \binits#1{#1}\fi
\ifx \bauthor  \undefined \def \bauthor#1{#1}\fi
\ifx \batitle  \undefined \def \batitle#1{#1}\fi
\ifx \bjtitle  \undefined \def \bjtitle#1{#1}\fi
\ifx \bvolume  \undefined \def \bvolume#1{\textbf{#1}}\fi
\ifx \byear  \undefined \def \byear#1{#1}\fi
\ifx \bissue  \undefined \def \bissue#1{#1}\fi
\ifx \bfpage  \undefined \def \bfpage#1{#1}\fi
\ifx \blpage  \undefined \def \blpage #1{#1}\fi
\ifx \burl  \undefined \def \burl#1{\textsf{#1}}\fi
\ifx \doiurl  \undefined \def \doiurl#1{\textsf{#1}}\fi
\ifx \betal  \undefined \def \betal{\textit{et al.}}\fi
\ifx \binstitute  \undefined \def \binstitute#1{#1}\fi
\ifx \binstitutionaled  \undefined \def \binstitutionaled#1{#1}\fi
\ifx \bctitle  \undefined \def \bctitle#1{#1}\fi
\ifx \beditor  \undefined \def \beditor#1{#1}\fi
\ifx \bpublisher  \undefined \def \bpublisher#1{#1}\fi
\ifx \bbtitle  \undefined \def \bbtitle#1{#1}\fi
\ifx \bedition  \undefined \def \bedition#1{#1}\fi
\ifx \bseriesno  \undefined \def \bseriesno#1{#1}\fi
\ifx \blocation  \undefined \def \blocation#1{#1}\fi
\ifx \bsertitle  \undefined \def \bsertitle#1{#1}\fi
\ifx \bsnm \undefined \def \bsnm#1{#1}\fi
\ifx \bsuffix \undefined \def \bsuffix#1{#1}\fi
\ifx \bparticle \undefined \def \bparticle#1{#1}\fi
\ifx \barticle \undefined \def \barticle#1{#1}\fi
\ifx \bconfdate \undefined \def \bconfdate #1{#1}\fi
\ifx \botherref \undefined \def \botherref #1{#1}\fi
\ifx \url \undefined \def \url#1{\textsf{#1}}\fi
\ifx \bchapter \undefined \def \bchapter#1{#1}\fi
\ifx \bbook \undefined \def \bbook#1{#1}\fi
\ifx \bcomment \undefined \def \bcomment#1{#1}\fi
\ifx \oauthor \undefined \def \oauthor#1{#1}\fi
\ifx \citeauthoryear \undefined \def \citeauthoryear#1{#1}\fi
\ifx \endbibitem  \undefined \def \endbibitem {}\fi
\ifx \bconflocation  \undefined \def \bconflocation#1{#1}\fi
\ifx \arxivurl  \undefined \def \arxivurl#1{\textsf{#1}}\fi
\csname PreBibitemsHook\endcsname

\bibitem{world2018global}
\begin{botherref}
\oauthor{\bsnm{Organization}, \binits{W.H.}}, et al.:
Global status report on road safety 2018: Summary.
Technical report,
World Health Organization
(2018)
\end{botherref}
\endbibitem

\bibitem{tingvall1999vision}
\begin{bchapter}
\bauthor{\bsnm{Tingvall}, \binits{C.}},
\bauthor{\bsnm{Haworth}, \binits{N.}}:
\bctitle{Vision zero-an ethical approach to safety and mobility}.
In: \bbtitle{6th ITE International Conference Road Safety \& Traffic
  Enforcement: Beyond 2000}
(\byear{1999})
\end{bchapter}
\endbibitem

\bibitem{kandt2021smart}
\begin{barticle}
\bauthor{\bsnm{Kandt}, \binits{J.}},
\bauthor{\bsnm{Batty}, \binits{M.}}:
\batitle{Smart cities, big data and urban policy: Towards urban analytics for
  the long run}.
\bjtitle{Cities}
\bvolume{109},
\bfpage{102992}
(\byear{2021})
\end{barticle}
\endbibitem

\bibitem{ibrahim2020understanding}
\begin{barticle}
\bauthor{\bsnm{Ibrahim}, \binits{M.R.}},
\bauthor{\bsnm{Haworth}, \binits{J.}},
\bauthor{\bsnm{Cheng}, \binits{T.}}:
\batitle{Understanding cities with machine eyes: A review of deep computer
  vision in urban analytics}.
\bjtitle{Cities}
\bvolume{96},
\bfpage{102481}
(\byear{2020})
\end{barticle}
\endbibitem

\bibitem{he2021inferring}
\begin{bchapter}
\bauthor{\bsnm{He}, \binits{S.}},
\bauthor{\bsnm{Sadeghi}, \binits{M.A.}},
\bauthor{\bsnm{Chawla}, \binits{S.}},
\bauthor{\bsnm{Alizadeh}, \binits{M.}},
\bauthor{\bsnm{Balakrishnan}, \binits{H.}},
\bauthor{\bsnm{Madden}, \binits{S.}}:
\bctitle{Inferring high-resolution traffic accident risk maps based on
  satellite imagery and gps trajectories}.
In: \bbtitle{Proceedings of the IEEE/CVF International Conference on Computer
  Vision},
pp. \bfpage{11977}--\blpage{11985}
(\byear{2021})
\end{bchapter}
\endbibitem

\bibitem{yoshimura2021street}
\begin{botherref}
\oauthor{\bsnm{Yoshimura}, \binits{Y.}},
\oauthor{\bsnm{Kumakoshi}, \binits{Y.}},
\oauthor{\bsnm{Fan}, \binits{Y.}},
\oauthor{\bsnm{Milardo}, \binits{S.}},
\oauthor{\bsnm{Koizumi}, \binits{H.}},
\oauthor{\bsnm{Santi}, \binits{P.}},
\oauthor{\bsnm{Arias}, \binits{J.M.}},
\oauthor{\bsnm{Zheng}, \binits{S.}},
\oauthor{\bsnm{Ratti}, \binits{C.}}:
Street pedestrianization in urban districts: Economic impacts in spanish
  cities.
Cities,
103468
(2021)
\end{botherref}
\endbibitem

\bibitem{naik2017computer}
\begin{barticle}
\bauthor{\bsnm{Naik}, \binits{N.}},
\bauthor{\bsnm{Kominers}, \binits{S.D.}},
\bauthor{\bsnm{Raskar}, \binits{R.}},
\bauthor{\bsnm{Glaeser}, \binits{E.L.}},
\bauthor{\bsnm{Hidalgo}, \binits{C.A.}}:
\batitle{Computer vision uncovers predictors of physical urban change}.
\bjtitle{Proceedings of the National Academy of Sciences}
\bvolume{114}(\bissue{29}),
\bfpage{7571}--\blpage{7576}
(\byear{2017})
\end{barticle}
\endbibitem

\bibitem{alhasoun2019streetify}
\begin{bchapter}
\bauthor{\bsnm{Alhasoun}, \binits{F.}},
\bauthor{\bsnm{Gonz{\'a}lez}, \binits{M.}}:
\bctitle{Streetify: Using street view imagery and deep learning for urban
  streets development}.
In: \bbtitle{2019 IEEE International Conference on Big Data (Big Data)},
pp. \bfpage{2001}--\blpage{2006}
(\byear{2019}).
\bcomment{IEEE}
\end{bchapter}
\endbibitem

\bibitem{song2018farsa}
\begin{bchapter}
\bauthor{\bsnm{Song}, \binits{W.}},
\bauthor{\bsnm{Workman}, \binits{S.}},
\bauthor{\bsnm{Hadzic}, \binits{A.}},
\bauthor{\bsnm{Zhang}, \binits{X.}},
\bauthor{\bsnm{Green}, \binits{E.}},
\bauthor{\bsnm{Chen}, \binits{M.}},
\bauthor{\bsnm{Souleyrette}, \binits{R.}},
\bauthor{\bsnm{Jacobs}, \binits{N.}}:
\bctitle{Farsa: Fully automated roadway safety assessment}.
In: \bbtitle{2018 IEEE Winter Conference on Applications of Computer Vision
  (WACV)},
pp. \bfpage{521}--\blpage{529}
(\byear{2018}).
\bcomment{IEEE}
\end{bchapter}
\endbibitem

\bibitem{bustos2021explainable}
\begin{barticle}
\bauthor{\bsnm{Bustos}, \binits{C.}},
\bauthor{\bsnm{Rhoads}, \binits{D.}},
\bauthor{\bsnm{Solé-Ribalta}, \binits{A.}},
\bauthor{\bsnm{Masip}, \binits{D.}},
\bauthor{\bsnm{Arenas}, \binits{A.}},
\bauthor{\bsnm{Lapedriza}, \binits{A.}},
\bauthor{\bsnm{Borge-Holthoefer}, \binits{J.}}:
\batitle{Explainable, automated urban interventions to improve pedestrian and
  vehicle safety}.
\bjtitle{Transportation Research Part C: Emerging Technologies}
\bvolume{125},
\bfpage{103018}
(\byear{2021}).
doi:\doiurl{10.1016/j.trc.2021.103018}
\end{barticle}
\endbibitem

\bibitem{kauer2018mapping}
\begin{barticle}
\bauthor{\bsnm{Kauer}, \binits{T.}},
\bauthor{\bsnm{Joglekar}, \binits{S.}},
\bauthor{\bsnm{Redi}, \binits{M.}},
\bauthor{\bsnm{Aiello}, \binits{L.M.}},
\bauthor{\bsnm{Quercia}, \binits{D.}}:
\batitle{Mapping and visualizing deep-learning urban beautification}.
\bjtitle{IEEE Computer Graphics and Applications}
\bvolume{38}(\bissue{5}),
\bfpage{70}--\blpage{83}
(\byear{2018})
\end{barticle}
\endbibitem

\bibitem{simini2021deep}
\begin{barticle}
\bauthor{\bsnm{Simini}, \binits{F.}},
\bauthor{\bsnm{Barlacchi}, \binits{G.}},
\bauthor{\bsnm{Luca}, \binits{M.}},
\bauthor{\bsnm{Pappalardo}, \binits{L.}}:
\batitle{A deep gravity model for mobility flows generation}.
\bjtitle{Nature communications}
\bvolume{12}(\bissue{1}),
\bfpage{1}--\blpage{13}
(\byear{2021})
\end{barticle}
\endbibitem

\bibitem{dubey2016deep}
\begin{bchapter}
\bauthor{\bsnm{Dubey}, \binits{A.}},
\bauthor{\bsnm{Naik}, \binits{N.}},
\bauthor{\bsnm{Parikh}, \binits{D.}},
\bauthor{\bsnm{Raskar}, \binits{R.}},
\bauthor{\bsnm{Hidalgo}, \binits{C.A.}}:
\bctitle{Deep learning the city: Quantifying urban perception at a global
  scale}.
In: \bbtitle{European Conference on Computer Vision},
pp. \bfpage{196}--\blpage{212}
(\byear{2016}).
\bcomment{Springer}
\end{bchapter}
\endbibitem

\bibitem{wu2021influence}
\begin{botherref}
\oauthor{\bsnm{Wu}, \binits{P.}},
\oauthor{\bsnm{Song}, \binits{L.}},
\oauthor{\bsnm{Meng}, \binits{X.}}:
Influence of built environment and roadway characteristics on the frequency of
  vehicle crashes caused by driver inattention: A comparison between rural
  roads and urban roads.
Journal of Safety Research
(2021)
\end{botherref}
\endbibitem

\bibitem{nasar2008mobile}
\begin{barticle}
\bauthor{\bsnm{Nasar}, \binits{J.}},
\bauthor{\bsnm{Hecht}, \binits{P.}},
\bauthor{\bsnm{Wener}, \binits{R.}}:
\batitle{Mobile telephones, distracted attention, and pedestrian safety}.
\bjtitle{Accident analysis \& prevention}
\bvolume{40}(\bissue{1}),
\bfpage{69}--\blpage{75}
(\byear{2008})
\end{barticle}
\endbibitem

\bibitem{rifaat2011effect}
\begin{barticle}
\bauthor{\bsnm{Rifaat}, \binits{S.M.}},
\bauthor{\bsnm{Tay}, \binits{R.}},
\bauthor{\bsnm{De~Barros}, \binits{A.}}:
\batitle{Effect of street pattern on the severity of crashes involving
  vulnerable road users}.
\bjtitle{Accident Analysis \& Prevention}
\bvolume{43}(\bissue{1}),
\bfpage{276}--\blpage{283}
(\byear{2011})
\end{barticle}
\endbibitem

\bibitem{moeinaddini2014relationship}
\begin{barticle}
\bauthor{\bsnm{Moeinaddini}, \binits{M.}},
\bauthor{\bsnm{Asadi-Shekari}, \binits{Z.}},
\bauthor{\bsnm{Shah}, \binits{M.Z.}}:
\batitle{The relationship between urban street networks and the number of
  transport fatalities at the city level}.
\bjtitle{Safety Science}
\bvolume{62},
\bfpage{114}--\blpage{120}
(\byear{2014})
\end{barticle}
\endbibitem

\bibitem{daraei2021data}
\begin{barticle}
\bauthor{\bsnm{Daraei}, \binits{S.}},
\bauthor{\bsnm{Pelechrinis}, \binits{K.}},
\bauthor{\bsnm{Quercia}, \binits{D.}}:
\batitle{A data-driven approach for assessing biking safety in cities}.
\bjtitle{EPJ Data Science}
\bvolume{10}(\bissue{1}),
\bfpage{1}--\blpage{16}
(\byear{2021})
\end{barticle}
\endbibitem

\bibitem{mukoko2019examining}
\begin{botherref}
\oauthor{\bsnm{Mukoko}, \binits{K.K.}},
\oauthor{\bsnm{Pulugurtha}, \binits{S.S.}}:
Examining the influence of network, land use, and demographic characteristics
  to estimate the number of bicycle-vehicle crashes on urban roads.
IATSS Research
(2019)
\end{botherref}
\endbibitem

\bibitem{mecredy2012neighbourhood}
\begin{barticle}
\bauthor{\bsnm{Mecredy}, \binits{G.}},
\bauthor{\bsnm{Janssen}, \binits{I.}},
\bauthor{\bsnm{Pickett}, \binits{W.}}:
\batitle{Neighbourhood street connectivity and injury in youth: a national
  study of built environments in canada}.
\bjtitle{Injury Prevention}
\bvolume{18}(\bissue{2}),
\bfpage{81}--\blpage{87}
(\byear{2012})
\end{barticle}
\endbibitem

\bibitem{fu2019investigating}
\begin{barticle}
\bauthor{\bsnm{Fu}, \binits{T.}},
\bauthor{\bsnm{Hu}, \binits{W.}},
\bauthor{\bsnm{Miranda-Moreno}, \binits{L.}},
\bauthor{\bsnm{Saunier}, \binits{N.}}:
\batitle{Investigating secondary pedestrian-vehicle interactions at
  non-signalized intersections using vision-based trajectory data}.
\bjtitle{Transportation Research Part C: Emerging Technologies}
\bvolume{105},
\bfpage{222}--\blpage{240}
(\byear{2019})
\end{barticle}
\endbibitem

\bibitem{hu2018dangerous}
\begin{barticle}
\bauthor{\bsnm{Hu}, \binits{Y.}},
\bauthor{\bsnm{Zhang}, \binits{Y.}},
\bauthor{\bsnm{Shelton}, \binits{K.S.}}:
\batitle{Where are the dangerous intersections for pedestrians and cyclists: A
  colocation-based approach}.
\bjtitle{Transportation Research Part C: Emerging Technologies}
\bvolume{95},
\bfpage{431}--\blpage{441}
(\byear{2018})
\end{barticle}
\endbibitem

\bibitem{ukkusuri2012role}
\begin{barticle}
\bauthor{\bsnm{Ukkusuri}, \binits{S.}},
\bauthor{\bsnm{Miranda-Moreno}, \binits{L.F.}},
\bauthor{\bsnm{Ramadurai}, \binits{G.}},
\bauthor{\bsnm{Isa-Tavarez}, \binits{J.}}:
\batitle{The role of built environment on pedestrian crash frequency}.
\bjtitle{Safety Science}
\bvolume{50}(\bissue{4}),
\bfpage{1141}--\blpage{1151}
(\byear{2012})
\end{barticle}
\endbibitem

\bibitem{zhang2020prediction}
\begin{barticle}
\bauthor{\bsnm{Zhang}, \binits{S.}},
\bauthor{\bsnm{Abdel-Aty}, \binits{M.}},
\bauthor{\bsnm{Yuan}, \binits{J.}},
\bauthor{\bsnm{Li}, \binits{P.}}:
\batitle{Prediction of pedestrian crossing intentions at intersections based on
  long short-term memory recurrent neural network}.
\bjtitle{Transportation research record}
\bvolume{2674}(\bissue{4}),
\bfpage{57}--\blpage{65}
(\byear{2020})
\end{barticle}
\endbibitem

\bibitem{chen2016effects}
\begin{barticle}
\bauthor{\bsnm{Chen}, \binits{P.}},
\bauthor{\bsnm{Zhou}, \binits{J.}}:
\batitle{Effects of the built environment on automobile-involved pedestrian
  crash frequency and risk}.
\bjtitle{Journal of Transport \& Health}
\bvolume{3}(\bissue{4}),
\bfpage{448}--\blpage{456}
(\byear{2016})
\end{barticle}
\endbibitem

\bibitem{palazzi2018predicting}
\begin{barticle}
\bauthor{\bsnm{Palazzi}, \binits{A.}},
\bauthor{\bsnm{Abati}, \binits{D.}},
\bauthor{\bsnm{Solera}, \binits{F.}},
\bauthor{\bsnm{Cucchiara}, \binits{R.}}, \betal:
\batitle{Predicting the driver's focus of attention: the dr(eye)ve project}.
\bjtitle{IEEE Transactions on Pattern Analysis and Machine Intelligence}
\bvolume{41}(\bissue{7}),
\bfpage{1720}--\blpage{1733}
(\byear{2018})
\end{barticle}
\endbibitem

\bibitem{khaki2021probabilistic}
\begin{botherref}
\oauthor{\bsnm{Khaki}, \binits{A.M.}},
\oauthor{\bsnm{Mohammadnazar}, \binits{A.}}:
A probabilistic model for pedestrian gap acceptance behavior at uncontrolled
  midblock crosswalks.
arXiv preprint arXiv:2101.10903
(2021)
\end{botherref}
\endbibitem

\bibitem{sargoni2020sequential}
\begin{bchapter}
\bauthor{\bsnm{Sargoni}, \binits{O.T.}},
\bauthor{\bsnm{Manley}, \binits{E.}}:
\bctitle{A sequential sampling model of pedestrian road crossing choice}.
In: \bbtitle{Proceedings of the 3rd ACM SIGSPATIAL International Workshop on
  GeoSpatial Simulation},
pp. \bfpage{10}--\blpage{19}
(\byear{2020})
\end{bchapter}
\endbibitem

\bibitem{bogacz2021modelling}
\begin{barticle}
\bauthor{\bsnm{Bogacz}, \binits{M.}},
\bauthor{\bsnm{Hess}, \binits{S.}},
\bauthor{\bsnm{Calastri}, \binits{C.}},
\bauthor{\bsnm{Choudhury}, \binits{C.F.}},
\bauthor{\bsnm{Mushtaq}, \binits{F.}},
\bauthor{\bsnm{Awais}, \binits{M.}},
\bauthor{\bsnm{Nazemi}, \binits{M.}},
\bauthor{\bparticle{van} \bsnm{Eggermond}, \binits{M.A.}},
\bauthor{\bsnm{Erath}, \binits{A.}}:
\batitle{Modelling risk perception using a dynamic hybrid choice model and
  brain-imaging data: Application to virtual reality cycling}.
\bjtitle{Transportation Research Part C: Emerging Technologies}
\bvolume{133},
\bfpage{103435}
(\byear{2021})
\end{barticle}
\endbibitem

\bibitem{alhazzani2021urban}
\begin{barticle}
\bauthor{\bsnm{Alhazzani}, \binits{M.}},
\bauthor{\bsnm{Alhasoun}, \binits{F.}},
\bauthor{\bsnm{Alawwad}, \binits{Z.}},
\bauthor{\bsnm{Gonz{\'a}lez}, \binits{M.C.}}:
\batitle{Urban attractors: Discovering patterns in regions of attraction in
  cities}.
\bjtitle{Plos one}
\bvolume{16}(\bissue{4}),
\bfpage{0250204}
(\byear{2021})
\end{barticle}
\endbibitem

\bibitem{anguelov2010google}
\begin{barticle}
\bauthor{\bsnm{Anguelov}, \binits{D.}},
\bauthor{\bsnm{Dulong}, \binits{C.}},
\bauthor{\bsnm{Filip}, \binits{D.}},
\bauthor{\bsnm{Frueh}, \binits{C.}},
\bauthor{\bsnm{Lafon}, \binits{S.}},
\bauthor{\bsnm{Lyon}, \binits{R.}},
\bauthor{\bsnm{Ogale}, \binits{A.}},
\bauthor{\bsnm{Vincent}, \binits{L.}},
\bauthor{\bsnm{Weaver}, \binits{J.}}:
\batitle{Google street view: Capturing the world at street level}.
\bjtitle{Computer}
\bvolume{43}(\bissue{6}),
\bfpage{32}--\blpage{38}
(\byear{2010})
\end{barticle}
\endbibitem

\bibitem{mapillary2019}
\begin{botherref}
\oauthor{\bsnm{{Mapillary contributors}}}:
{Mapillary - Street-level imagery, powered by collaboration and computer vision
  }.
\url{ https://www.mapillary.com/app }
(2019)
\end{botherref}
\endbibitem

\bibitem{moray1959attention}
\begin{barticle}
\bauthor{\bsnm{Moray}, \binits{N.}}:
\batitle{Attention in dichotic listening: Affective cues and the influence of
  instructions}.
\bjtitle{Quarterly journal of experimental psychology}
\bvolume{11}(\bissue{1}),
\bfpage{56}--\blpage{60}
(\byear{1959})
\end{barticle}
\endbibitem

\bibitem{kahneman1973attention}
\begin{bbook}
\bauthor{\bsnm{Kahneman}, \binits{D.}}:
\bbtitle{Attention and Effort}
vol. \bseriesno{1063}.
\bpublisher{Citeseer}, \blocation{???}
(\byear{1973})
\end{bbook}
\endbibitem

\bibitem{alvarez2004capacity}
\begin{barticle}
\bauthor{\bsnm{Alvarez}, \binits{G.A.}},
\bauthor{\bsnm{Cavanagh}, \binits{P.}}:
\batitle{The capacity of visual short-term memory is set both by visual
  information load and by number of objects}.
\bjtitle{Psychological science}
\bvolume{15}(\bissue{2}),
\bfpage{106}--\blpage{111}
(\byear{2004})
\end{barticle}
\endbibitem

\bibitem{richards2010development}
\begin{barticle}
\bauthor{\bsnm{Richards}, \binits{J.E.}}:
\batitle{The development of attention to simple and complex visual stimuli in
  infants: Behavioral and psychophysiological measures}.
\bjtitle{Developmental Review}
\bvolume{30}(\bissue{2}),
\bfpage{203}--\blpage{219}
(\byear{2010})
\end{barticle}
\endbibitem

\bibitem{bustos2021stress}
\begin{bchapter}
\bauthor{\bsnm{Bustos}, \binits{C.}},
\bauthor{\bsnm{Elhaouij}, \binits{N.}},
\bauthor{\bsnm{Sole-Ribalta}, \binits{A.}},
\bauthor{\bsnm{Borge-Holthoefer}, \binits{J.}},
\bauthor{\bsnm{Lapedriza}, \binits{A.}},
\bauthor{\bsnm{Picard}, \binits{R.}}:
\bctitle{Predicting driver self-reported stress by analyzing the road scene}.
In: \bbtitle{2021 9th International Conference on Affective Computing and
  Intelligent Interaction (ACII)},
p.
(\byear{2021}).
\bcomment{IEEE}
\end{bchapter}
\endbibitem

\bibitem{ito2021assessing}
\begin{botherref}
\oauthor{\bsnm{Ito}, \binits{K.}},
\oauthor{\bsnm{Biljecki}, \binits{F.}}:
Assessing bikeability with street view imagery and computer vision.
arXiv preprint arXiv:2105.08499
(2021)
\end{botherref}
\endbibitem

\bibitem{lu2019using}
\begin{barticle}
\bauthor{\bsnm{Lu}, \binits{Y.}}:
\batitle{Using google street view to investigate the association between street
  greenery and physical activity}.
\bjtitle{Landscape and Urban Planning}
\bvolume{191},
\bfpage{103435}
(\byear{2019})
\end{barticle}
\endbibitem

\bibitem{kita2019google}
\begin{botherref}
\oauthor{\bsnm{Kita}, \binits{K.}},
\oauthor{\bsnm{Kidzi{\'n}ski}, \binits{{\L}.}}:
Google street view image of a house predicts car accident risk of its resident.
arXiv preprint arXiv:1904.05270
(2019)
\end{botherref}
\endbibitem

\bibitem{naik2014streetscore}
\begin{bchapter}
\bauthor{\bsnm{Naik}, \binits{N.}},
\bauthor{\bsnm{Philipoom}, \binits{J.}},
\bauthor{\bsnm{Raskar}, \binits{R.}},
\bauthor{\bsnm{Hidalgo}, \binits{C.}}:
\bctitle{Streetscore-predicting the perceived safety of one million
  streetscapes}.
In: \bbtitle{Proceedings of the IEEE Conference on Computer Vision and Pattern
  Recognition Workshops},
pp. \bfpage{779}--\blpage{785}
(\byear{2014})
\end{bchapter}
\endbibitem

\bibitem{gebru2017pnas}
\begin{barticle}
\bauthor{\bsnm{Gebru}, \binits{T.}},
\bauthor{\bsnm{Krause}, \binits{J.}},
\bauthor{\bsnm{Wang}, \binits{Y.}},
\bauthor{\bsnm{Chen}, \binits{D.}},
\bauthor{\bsnm{Deng}, \binits{J.}},
\bauthor{\bsnm{Aiden}, \binits{E.L.}},
\bauthor{\bsnm{Fei-Fei}, \binits{L.}}:
\batitle{Using deep learning and google street view to estimate the demographic
  makeup of neighborhoods across the united states}.
\bjtitle{Proceedings of the National Academy of Sciences}
\bvolume{114}(\bissue{50}),
\bfpage{13108}--\blpage{13113}
(\byear{2017})
\end{barticle}
\endbibitem

\bibitem{xu2020towards}
\begin{botherref}
\oauthor{\bsnm{Xu}, \binits{R.}},
\oauthor{\bsnm{Lin}, \binits{A.Y.}},
\oauthor{\bsnm{Zhang}, \binits{S.}},
\oauthor{\bsnm{Xiong}, \binits{P.}},
\oauthor{\bsnm{Hecht}, \binits{B.}}:
Towards better driver safety: Empowering personal navigation technologies with
  road safety awareness.
arXiv preprint arXiv:2006.03196
(2020)
\end{botherref}
\endbibitem

\bibitem{rhoads2020planning}
\begin{botherref}
\oauthor{\bsnm{Rhoads}, \binits{D.}},
\oauthor{\bsnm{Sol{\'e}-Ribalta}, \binits{A.}},
\oauthor{\bsnm{Gonz{\'a}lez}, \binits{M.C.}},
\oauthor{\bsnm{Borge-Holthoefer}, \binits{J.}}:
Planning for sustainable open streets in pandemic cities.
arXiv preprint arXiv:2009.12548
(2020)
\end{botherref}
\endbibitem

\bibitem{de2015personalized}
\begin{barticle}
\bauthor{\bsnm{De~Domenico}, \binits{M.}},
\bauthor{\bsnm{Lima}, \binits{A.}},
\bauthor{\bsnm{Gonz{\'a}lez}, \binits{M.C.}},
\bauthor{\bsnm{Arenas}, \binits{A.}}:
\batitle{Personalized routing for multitudes in smart cities}.
\bjtitle{EPJ Data Science}
\bvolume{4}(\bissue{1}),
\bfpage{1}--\blpage{11}
(\byear{2015})
\end{barticle}
\endbibitem

\bibitem{bcn2019acc}
\begin{botherref}
\oauthor{\bsnm{{Ajuntament de Barcelona}}}:
Open Data BCN.
\url{https://opendata-ajuntament.barcelona.cat/en/}.
Accessed: 2019-04-20
(2019)
\end{botherref}
\endbibitem

\bibitem{OpenStreetMap}
\begin{botherref}
\oauthor{\bsnm{{OpenStreetMap contributors}}}:
{Planet dump retrieved from https://planet.osm.org }.
\url{ https://www.openstreetmap.org }
(2017)
\end{botherref}
\endbibitem

\bibitem{resnet}
\begin{bchapter}
\bauthor{\bsnm{He}, \binits{K.}},
\bauthor{\bsnm{Zhang}, \binits{X.}},
\bauthor{\bsnm{Ren}, \binits{S.}},
\bauthor{\bsnm{Sun}, \binits{J.}}:
\bctitle{Deep residual learning for image recognition}.
In: \bbtitle{Proceedings of the IEEE Conference on Computer Vision and Pattern
  Recognition},
pp. \bfpage{770}--\blpage{778}
(\byear{2016})
\end{bchapter}
\endbibitem

\bibitem{rhoads2021sustainable}
\begin{barticle}
\bauthor{\bsnm{Rhoads}, \binits{D.}},
\bauthor{\bsnm{Sol{\'e}-Ribalta}, \binits{A.}},
\bauthor{\bsnm{Gonz{\'a}lez}, \binits{M.C.}},
\bauthor{\bsnm{Borge-Holthoefer}, \binits{J.}}:
\batitle{A sustainable strategy for open streets in (post) pandemic cities}.
\bjtitle{Communications Physics}
\bvolume{4}(\bissue{1}),
\bfpage{1}--\blpage{12}
(\byear{2021})
\end{barticle}
\endbibitem

\bibitem{yang2014limits}
\begin{barticle}
\bauthor{\bsnm{Yang}, \binits{Y.}},
\bauthor{\bsnm{Herrera}, \binits{C.}},
\bauthor{\bsnm{Eagle}, \binits{N.}},
\bauthor{\bsnm{Gonz{\'a}lez}, \binits{M.C.}}:
\batitle{Limits of predictability in commuting flows in the absence of data for
  calibration}.
\bjtitle{Scientific reports}
\bvolume{4}(\bissue{1}),
\bfpage{1}--\blpage{9}
(\byear{2014})
\end{barticle}
\endbibitem

\bibitem{yang2016participatory}
\begin{barticle}
\bauthor{\bsnm{Yang}, \binits{D.}},
\bauthor{\bsnm{Zhang}, \binits{D.}},
\bauthor{\bsnm{Qu}, \binits{B.}}:
\batitle{Participatory cultural mapping based on collective behavior data in
  location-based social networks}.
\bjtitle{ACM Transactions on Intelligent Systems and Technology (TIST)}
\bvolume{7}(\bissue{3}),
\bfpage{30}
(\byear{2016})
\end{barticle}
\endbibitem

\bibitem{yang2015nationtelescope}
\begin{barticle}
\bauthor{\bsnm{Yang}, \binits{D.}},
\bauthor{\bsnm{Zhang}, \binits{D.}},
\bauthor{\bsnm{Chen}, \binits{L.}},
\bauthor{\bsnm{Qu}, \binits{B.}}:
\batitle{Nationtelescope: Monitoring and visualizing large-scale collective
  behavior in lbsns}.
\bjtitle{Journal of Network and Computer Applications}
\bvolume{55},
\bfpage{170}--\blpage{180}
(\byear{2015})
\end{barticle}
\endbibitem

\bibitem{sole2018decongestion}
\begin{barticle}
\bauthor{\bsnm{Sol{\'e}-Ribalta}, \binits{A.}},
\bauthor{\bsnm{G{\'o}mez}, \binits{S.}},
\bauthor{\bsnm{Arenas}, \binits{A.}}:
\batitle{Decongestion of urban areas with hotspot pricing}.
\bjtitle{Networks and Spatial Economics}
\bvolume{18}(\bissue{1}),
\bfpage{33}--\blpage{50}
(\byear{2018})
\end{barticle}
\endbibitem

\bibitem{henry2019spatio}
\begin{bchapter}
\bauthor{\bsnm{Henry}, \binits{E.}},
\bauthor{\bsnm{Bonnetain}, \binits{L.}},
\bauthor{\bsnm{Furno}, \binits{A.}},
\bauthor{\bsnm{El~Faouzi}, \binits{N.-E.}},
\bauthor{\bsnm{Zimeo}, \binits{E.}}:
\bctitle{Spatio-temporal correlations of betweenness centrality and traffic
  metrics}.
In: \bbtitle{2019 6th International Conference on Models and Technologies for
  Intelligent Transportation Systems (MT-ITS)},
pp. \bfpage{1}--\blpage{10}
(\byear{2019}).
\bcomment{IEEE}
\end{bchapter}
\endbibitem

\bibitem{altshuler2011augmented}
\begin{bchapter}
\bauthor{\bsnm{Altshuler}, \binits{Y.}},
\bauthor{\bsnm{Puzis}, \binits{R.}},
\bauthor{\bsnm{Elovici}, \binits{Y.}},
\bauthor{\bsnm{Bekhor}, \binits{S.}},
\bauthor{\bsnm{Pentland}, \binits{A.}}:
\bctitle{Augmented betweenness centrality for mobility prediction in
  transportation networks}.
In: \bbtitle{International Workshop on Finding Patterns of Human Behaviors in
  NEtworks and MObility Data, NEMO11}
(\byear{2011})
\end{bchapter}
\endbibitem

\bibitem{bcnVisioCero}
\begin{botherref}
La Guardia Urbana comprometida con el objetivo Vision Cero
(2017)
\end{botherref}
\endbibitem

\bibitem{mueller2020changing}
\begin{barticle}
\bauthor{\bsnm{Mueller}, \binits{N.}},
\bauthor{\bsnm{Rojas-Rueda}, \binits{D.}},
\bauthor{\bsnm{Khreis}, \binits{H.}},
\bauthor{\bsnm{Cirach}, \binits{M.}},
\bauthor{\bsnm{Andr{\'e}s}, \binits{D.}},
\bauthor{\bsnm{Ballester}, \binits{J.}},
\bauthor{\bsnm{Bartoll}, \binits{X.}},
\bauthor{\bsnm{Daher}, \binits{C.}},
\bauthor{\bsnm{Deluca}, \binits{A.}},
\bauthor{\bsnm{Echave}, \binits{C.}}, \betal:
\batitle{Changing the urban design of cities for health: The superblock model}.
\bjtitle{Environment international}
\bvolume{134},
\bfpage{105132}
(\byear{2020})
\end{barticle}
\endbibitem

\bibitem{silva2016tactical}
\begin{barticle}
\bauthor{\bsnm{Silva}, \binits{P.}}:
\batitle{Tactical urbanism: Towards an evolutionary cities’ approach?}
\bjtitle{Environment and Planning B: Planning and design}
\bvolume{43}(\bissue{6}),
\bfpage{1040}--\blpage{1051}
(\byear{2016})
\end{barticle}
\endbibitem

\bibitem{newman2009networks}
\begin{bbook}
\bauthor{\bsnm{Newman}, \binits{M.}}:
\bbtitle{Networks: An Introduction}.
\bpublisher{OUP Oxford}, \blocation{???}
(\byear{2009})
\end{bbook}
\endbibitem

\bibitem{freeman1977set}
\begin{botherref}
\oauthor{\bsnm{Freeman}, \binits{L.C.}}:
A set of measures of centrality based on betweenness.
Sociometry,
35--41
(1977)
\end{botherref}
\endbibitem

\bibitem{mahfouz2021road}
\begin{botherref}
\oauthor{\bsnm{Mahfouz}, \binits{H.}},
\oauthor{\bsnm{Arcaute}, \binits{E.}},
\oauthor{\bsnm{Lovelace}, \binits{R.}}:
A road segment prioritization approach for cycling infrastructure.
arXiv preprint arXiv:2105.03712
(2021)
\end{botherref}
\endbibitem

\bibitem{bulo2018place}
\begin{bchapter}
\bauthor{\bsnm{Bulo}, \binits{S.R.}},
\bauthor{\bsnm{Porzi}, \binits{L.}},
\bauthor{\bsnm{Kontschieder}, \binits{P.}}:
\bctitle{In-place activated batchnorm for memory-optimized training of dnns}.
In: \bbtitle{Proceedings of the IEEE Conference on Computer Vision and Pattern
  Recognition},
pp. \bfpage{5639}--\blpage{5647}
(\byear{2018})
\end{bchapter}
\endbibitem

\bibitem{neuhold2017mapillary}
\begin{bchapter}
\bauthor{\bsnm{Neuhold}, \binits{G.}},
\bauthor{\bsnm{Ollmann}, \binits{T.}},
\bauthor{\bsnm{Rota~Bulo}, \binits{S.}},
\bauthor{\bsnm{Kontschieder}, \binits{P.}}:
\bctitle{The mapillary vistas dataset for semantic understanding of street
  scenes}.
In: \bbtitle{Proceedings of the IEEE International Conference on Computer
  Vision},
pp. \bfpage{4990}--\blpage{4999}
(\byear{2017})
\end{bchapter}
\endbibitem

\end{thebibliography}

\newcommand{\BMCxmlcomment}[1]{}

\BMCxmlcomment{

<refgrp>

<bibl id="B1">
  <title><p>Global status report on road safety 2018: Summary</p></title>
  <aug>
    <au><snm>Organization</snm><fnm>WH</fnm></au>
    <au><cnm>others</cnm></au>
  </aug>
  <pubdate>2018</pubdate>
</bibl>

<bibl id="B2">
  <title><p>Vision Zero-An ethical approach to safety and mobility</p></title>
  <aug>
    <au><snm>Tingvall</snm><fnm>C</fnm></au>
    <au><snm>Haworth</snm><fnm>N</fnm></au>
  </aug>
  <source>6th ITE International Conference Road Safety \& Traffic Enforcement:
  Beyond 2000</source>
  <pubdate>1999</pubdate>
</bibl>

<bibl id="B3">
  <title><p>Smart cities, big data and urban policy: Towards urban analytics
  for the long run</p></title>
  <aug>
    <au><snm>Kandt</snm><fnm>J</fnm></au>
    <au><snm>Batty</snm><fnm>M</fnm></au>
  </aug>
  <source>Cities</source>
  <publisher>Elsevier</publisher>
  <pubdate>2021</pubdate>
  <volume>109</volume>
  <fpage>102992</fpage>
</bibl>

<bibl id="B4">
  <title><p>Understanding cities with machine eyes: A review of deep computer
  vision in urban analytics</p></title>
  <aug>
    <au><snm>Ibrahim</snm><fnm>MR</fnm></au>
    <au><snm>Haworth</snm><fnm>J</fnm></au>
    <au><snm>Cheng</snm><fnm>T</fnm></au>
  </aug>
  <source>Cities</source>
  <publisher>Elsevier</publisher>
  <pubdate>2020</pubdate>
  <volume>96</volume>
  <fpage>102481</fpage>
</bibl>

<bibl id="B5">
  <title><p>Inferring high-resolution traffic accident risk maps based on
  satellite imagery and GPS trajectories</p></title>
  <aug>
    <au><snm>He</snm><fnm>S</fnm></au>
    <au><snm>Sadeghi</snm><fnm>MA</fnm></au>
    <au><snm>Chawla</snm><fnm>S</fnm></au>
    <au><snm>Alizadeh</snm><fnm>M</fnm></au>
    <au><snm>Balakrishnan</snm><fnm>H</fnm></au>
    <au><snm>Madden</snm><fnm>S</fnm></au>
  </aug>
  <source>Proceedings of the IEEE/CVF International Conference on Computer
  Vision</source>
  <pubdate>2021</pubdate>
  <fpage>11977</fpage>
  <lpage>-11985</lpage>
</bibl>

<bibl id="B6">
  <title><p>Street pedestrianization in urban districts: Economic impacts in
  Spanish cities</p></title>
  <aug>
    <au><snm>Yoshimura</snm><fnm>Y</fnm></au>
    <au><snm>Kumakoshi</snm><fnm>Y</fnm></au>
    <au><snm>Fan</snm><fnm>Y</fnm></au>
    <au><snm>Milardo</snm><fnm>S</fnm></au>
    <au><snm>Koizumi</snm><fnm>H</fnm></au>
    <au><snm>Santi</snm><fnm>P</fnm></au>
    <au><snm>Arias</snm><fnm>JM</fnm></au>
    <au><snm>Zheng</snm><fnm>S</fnm></au>
    <au><snm>Ratti</snm><fnm>C</fnm></au>
  </aug>
  <source>Cities</source>
  <publisher>Elsevier</publisher>
  <pubdate>2021</pubdate>
  <fpage>103468</fpage>
</bibl>

<bibl id="B7">
  <title><p>Computer vision uncovers predictors of physical urban
  change</p></title>
  <aug>
    <au><snm>Naik</snm><fnm>N</fnm></au>
    <au><snm>Kominers</snm><fnm>SD</fnm></au>
    <au><snm>Raskar</snm><fnm>R</fnm></au>
    <au><snm>Glaeser</snm><fnm>EL</fnm></au>
    <au><snm>Hidalgo</snm><fnm>CA</fnm></au>
  </aug>
  <source>Proceedings of the National Academy of Sciences</source>
  <publisher>National Acad Sciences</publisher>
  <pubdate>2017</pubdate>
  <volume>114</volume>
  <issue>29</issue>
  <fpage>7571</fpage>
  <lpage>-7576</lpage>
</bibl>

<bibl id="B8">
  <title><p>Streetify: Using Street View Imagery And Deep Learning For Urban
  Streets Development</p></title>
  <aug>
    <au><snm>Alhasoun</snm><fnm>F</fnm></au>
    <au><snm>Gonz{\'a}lez</snm><fnm>M</fnm></au>
  </aug>
  <source>2019 IEEE International Conference on Big Data (Big Data)</source>
  <pubdate>2019</pubdate>
  <fpage>2001</fpage>
  <lpage>-2006</lpage>
</bibl>

<bibl id="B9">
  <title><p>Farsa: Fully automated roadway safety assessment</p></title>
  <aug>
    <au><snm>Song</snm><fnm>W</fnm></au>
    <au><snm>Workman</snm><fnm>S</fnm></au>
    <au><snm>Hadzic</snm><fnm>A</fnm></au>
    <au><snm>Zhang</snm><fnm>X</fnm></au>
    <au><snm>Green</snm><fnm>E</fnm></au>
    <au><snm>Chen</snm><fnm>M</fnm></au>
    <au><snm>Souleyrette</snm><fnm>R</fnm></au>
    <au><snm>Jacobs</snm><fnm>N</fnm></au>
  </aug>
  <source>2018 IEEE Winter Conference on Applications of Computer Vision
  (WACV)</source>
  <pubdate>2018</pubdate>
  <fpage>521</fpage>
  <lpage>-529</lpage>
</bibl>

<bibl id="B10">
  <title><p>Explainable, automated urban interventions to improve pedestrian
  and vehicle safety</p></title>
  <aug>
    <au><snm>Bustos</snm><fnm>C.</fnm></au>
    <au><snm>Rhoads</snm><fnm>D.</fnm></au>
    <au><snm>Solé Ribalta</snm><fnm>A.</fnm></au>
    <au><snm>Masip</snm><fnm>D.</fnm></au>
    <au><snm>Arenas</snm><fnm>A.</fnm></au>
    <au><snm>Lapedriza</snm><fnm>A.</fnm></au>
    <au><snm>Borge Holthoefer</snm><fnm>J.</fnm></au>
  </aug>
  <source>Transportation Research Part C: Emerging Technologies</source>
  <pubdate>2021</pubdate>
  <volume>125</volume>
  <fpage>103018</fpage>
  <url>https://www.sciencedirect.com/science/article/pii/S0968090X21000498</url>
</bibl>

<bibl id="B11">
  <title><p>Mapping and Visualizing Deep-Learning Urban
  Beautification</p></title>
  <aug>
    <au><snm>Kauer</snm><fnm>T</fnm></au>
    <au><snm>Joglekar</snm><fnm>S</fnm></au>
    <au><snm>Redi</snm><fnm>M</fnm></au>
    <au><snm>Aiello</snm><fnm>LM</fnm></au>
    <au><snm>Quercia</snm><fnm>D</fnm></au>
  </aug>
  <source>IEEE Computer Graphics and Applications</source>
  <publisher>IEEE</publisher>
  <pubdate>2018</pubdate>
  <volume>38</volume>
  <issue>5</issue>
  <fpage>70</fpage>
  <lpage>-83</lpage>
</bibl>

<bibl id="B12">
  <title><p>A Deep Gravity model for mobility flows generation</p></title>
  <aug>
    <au><snm>Simini</snm><fnm>F</fnm></au>
    <au><snm>Barlacchi</snm><fnm>G</fnm></au>
    <au><snm>Luca</snm><fnm>M</fnm></au>
    <au><snm>Pappalardo</snm><fnm>L</fnm></au>
  </aug>
  <source>Nature communications</source>
  <publisher>Nature Publishing Group</publisher>
  <pubdate>2021</pubdate>
  <volume>12</volume>
  <issue>1</issue>
  <fpage>1</fpage>
  <lpage>-13</lpage>
</bibl>

<bibl id="B13">
  <title><p>Deep learning the city: Quantifying urban perception at a global
  scale</p></title>
  <aug>
    <au><snm>Dubey</snm><fnm>A</fnm></au>
    <au><snm>Naik</snm><fnm>N</fnm></au>
    <au><snm>Parikh</snm><fnm>D</fnm></au>
    <au><snm>Raskar</snm><fnm>R</fnm></au>
    <au><snm>Hidalgo</snm><fnm>CA</fnm></au>
  </aug>
  <source>European conference on computer vision</source>
  <pubdate>2016</pubdate>
  <fpage>196</fpage>
  <lpage>-212</lpage>
</bibl>

<bibl id="B14">
  <title><p>Influence of built environment and roadway characteristics on the
  frequency of vehicle crashes caused by driver inattention: A comparison
  between rural roads and urban roads</p></title>
  <aug>
    <au><snm>Wu</snm><fnm>P</fnm></au>
    <au><snm>Song</snm><fnm>L</fnm></au>
    <au><snm>Meng</snm><fnm>X</fnm></au>
  </aug>
  <source>Journal of Safety Research</source>
  <publisher>Elsevier</publisher>
  <pubdate>2021</pubdate>
</bibl>

<bibl id="B15">
  <title><p>Mobile telephones, distracted attention, and pedestrian
  safety</p></title>
  <aug>
    <au><snm>Nasar</snm><fnm>J</fnm></au>
    <au><snm>Hecht</snm><fnm>P</fnm></au>
    <au><snm>Wener</snm><fnm>R</fnm></au>
  </aug>
  <source>Accident analysis \& prevention</source>
  <publisher>Elsevier</publisher>
  <pubdate>2008</pubdate>
  <volume>40</volume>
  <issue>1</issue>
  <fpage>69</fpage>
  <lpage>-75</lpage>
</bibl>

<bibl id="B16">
  <title><p>Effect of street pattern on the severity of crashes involving
  vulnerable road users</p></title>
  <aug>
    <au><snm>Rifaat</snm><fnm>SM</fnm></au>
    <au><snm>Tay</snm><fnm>R</fnm></au>
    <au><snm>De Barros</snm><fnm>A</fnm></au>
  </aug>
  <source>Accident Analysis \& Prevention</source>
  <publisher>Elsevier</publisher>
  <pubdate>2011</pubdate>
  <volume>43</volume>
  <issue>1</issue>
  <fpage>276</fpage>
  <lpage>-283</lpage>
</bibl>

<bibl id="B17">
  <title><p>The relationship between urban street networks and the number of
  transport fatalities at the city level</p></title>
  <aug>
    <au><snm>Moeinaddini</snm><fnm>M</fnm></au>
    <au><snm>Asadi Shekari</snm><fnm>Z</fnm></au>
    <au><snm>Shah</snm><fnm>MZ</fnm></au>
  </aug>
  <source>Safety Science</source>
  <publisher>Elsevier</publisher>
  <pubdate>2014</pubdate>
  <volume>62</volume>
  <fpage>114</fpage>
  <lpage>-120</lpage>
</bibl>

<bibl id="B18">
  <title><p>A data-driven approach for assessing biking safety in
  cities</p></title>
  <aug>
    <au><snm>Daraei</snm><fnm>S</fnm></au>
    <au><snm>Pelechrinis</snm><fnm>K</fnm></au>
    <au><snm>Quercia</snm><fnm>D</fnm></au>
  </aug>
  <source>EPJ Data Science</source>
  <publisher>SpringerOpen</publisher>
  <pubdate>2021</pubdate>
  <volume>10</volume>
  <issue>1</issue>
  <fpage>1</fpage>
  <lpage>-16</lpage>
</bibl>

<bibl id="B19">
  <title><p>Examining the influence of network, land use, and demographic
  characteristics to estimate the number of bicycle-vehicle crashes on urban
  roads</p></title>
  <aug>
    <au><snm>Mukoko</snm><fnm>KK</fnm></au>
    <au><snm>Pulugurtha</snm><fnm>SS</fnm></au>
  </aug>
  <source>IATSS Research</source>
  <publisher>Elsevier</publisher>
  <pubdate>2019</pubdate>
</bibl>

<bibl id="B20">
  <title><p>Neighbourhood street connectivity and injury in youth: a national
  study of built environments in Canada</p></title>
  <aug>
    <au><snm>Mecredy</snm><fnm>G</fnm></au>
    <au><snm>Janssen</snm><fnm>I</fnm></au>
    <au><snm>Pickett</snm><fnm>W</fnm></au>
  </aug>
  <source>Injury Prevention</source>
  <publisher>BMJ Publishing Group Ltd</publisher>
  <pubdate>2012</pubdate>
  <volume>18</volume>
  <issue>2</issue>
  <fpage>81</fpage>
  <lpage>-87</lpage>
</bibl>

<bibl id="B21">
  <title><p>Investigating secondary pedestrian-vehicle interactions at
  non-signalized intersections using vision-based trajectory data</p></title>
  <aug>
    <au><snm>Fu</snm><fnm>T</fnm></au>
    <au><snm>Hu</snm><fnm>W</fnm></au>
    <au><snm>Miranda Moreno</snm><fnm>L</fnm></au>
    <au><snm>Saunier</snm><fnm>N</fnm></au>
  </aug>
  <source>Transportation Research Part C: Emerging Technologies</source>
  <publisher>Elsevier</publisher>
  <pubdate>2019</pubdate>
  <volume>105</volume>
  <fpage>222</fpage>
  <lpage>-240</lpage>
</bibl>

<bibl id="B22">
  <title><p>Where are the dangerous intersections for pedestrians and cyclists:
  A colocation-based approach</p></title>
  <aug>
    <au><snm>Hu</snm><fnm>Y</fnm></au>
    <au><snm>Zhang</snm><fnm>Y</fnm></au>
    <au><snm>Shelton</snm><fnm>KS</fnm></au>
  </aug>
  <source>Transportation Research Part C: Emerging Technologies</source>
  <publisher>Elsevier</publisher>
  <pubdate>2018</pubdate>
  <volume>95</volume>
  <fpage>431</fpage>
  <lpage>-441</lpage>
</bibl>

<bibl id="B23">
  <title><p>The role of built environment on pedestrian crash
  frequency</p></title>
  <aug>
    <au><snm>Ukkusuri</snm><fnm>S</fnm></au>
    <au><snm>Miranda Moreno</snm><fnm>LF</fnm></au>
    <au><snm>Ramadurai</snm><fnm>G</fnm></au>
    <au><snm>Isa Tavarez</snm><fnm>J</fnm></au>
  </aug>
  <source>Safety Science</source>
  <publisher>Elsevier</publisher>
  <pubdate>2012</pubdate>
  <volume>50</volume>
  <issue>4</issue>
  <fpage>1141</fpage>
  <lpage>-1151</lpage>
</bibl>

<bibl id="B24">
  <title><p>Prediction of pedestrian crossing intentions at intersections based
  on long short-term memory recurrent neural network</p></title>
  <aug>
    <au><snm>Zhang</snm><fnm>S</fnm></au>
    <au><snm>Abdel Aty</snm><fnm>M</fnm></au>
    <au><snm>Yuan</snm><fnm>J</fnm></au>
    <au><snm>Li</snm><fnm>P</fnm></au>
  </aug>
  <source>Transportation research record</source>
  <publisher>SAGE Publications Sage CA: Los Angeles, CA</publisher>
  <pubdate>2020</pubdate>
  <volume>2674</volume>
  <issue>4</issue>
  <fpage>57</fpage>
  <lpage>-65</lpage>
</bibl>

<bibl id="B25">
  <title><p>Effects of the built environment on automobile-involved pedestrian
  crash frequency and risk</p></title>
  <aug>
    <au><snm>Chen</snm><fnm>P</fnm></au>
    <au><snm>Zhou</snm><fnm>J</fnm></au>
  </aug>
  <source>Journal of Transport \& Health</source>
  <publisher>Elsevier</publisher>
  <pubdate>2016</pubdate>
  <volume>3</volume>
  <issue>4</issue>
  <fpage>448</fpage>
  <lpage>-456</lpage>
</bibl>

<bibl id="B26">
  <title><p>Predicting the Driver's Focus of Attention: the DR(eye)VE
  Project</p></title>
  <aug>
    <au><snm>Palazzi</snm><fnm>A</fnm></au>
    <au><snm>Abati</snm><fnm>D</fnm></au>
    <au><snm>Solera</snm><fnm>F</fnm></au>
    <au><snm>Cucchiara</snm><fnm>R</fnm></au>
    <au><cnm>others</cnm></au>
  </aug>
  <source>IEEE Transactions on Pattern Analysis and Machine
  Intelligence</source>
  <publisher>IEEE</publisher>
  <pubdate>2018</pubdate>
  <volume>41</volume>
  <issue>7</issue>
  <fpage>1720</fpage>
  <lpage>-1733</lpage>
</bibl>

<bibl id="B27">
  <title><p>A probabilistic model for pedestrian gap acceptance behavior at
  uncontrolled midblock crosswalks</p></title>
  <aug>
    <au><snm>Khaki</snm><fnm>AM</fnm></au>
    <au><snm>Mohammadnazar</snm><fnm>A</fnm></au>
  </aug>
  <source>arXiv preprint arXiv:2101.10903</source>
  <pubdate>2021</pubdate>
</bibl>

<bibl id="B28">
  <title><p>A sequential sampling model of pedestrian road crossing
  choice</p></title>
  <aug>
    <au><snm>Sargoni</snm><fnm>OT</fnm></au>
    <au><snm>Manley</snm><fnm>E</fnm></au>
  </aug>
  <source>Proceedings of the 3rd ACM SIGSPATIAL International Workshop on
  GeoSpatial Simulation</source>
  <pubdate>2020</pubdate>
  <fpage>10</fpage>
  <lpage>-19</lpage>
</bibl>

<bibl id="B29">
  <title><p>Modelling risk perception using a dynamic hybrid choice model and
  brain-imaging data: Application to virtual reality cycling</p></title>
  <aug>
    <au><snm>Bogacz</snm><fnm>M</fnm></au>
    <au><snm>Hess</snm><fnm>S</fnm></au>
    <au><snm>Calastri</snm><fnm>C</fnm></au>
    <au><snm>Choudhury</snm><fnm>CF</fnm></au>
    <au><snm>Mushtaq</snm><fnm>F</fnm></au>
    <au><snm>Awais</snm><fnm>M</fnm></au>
    <au><snm>Nazemi</snm><fnm>M</fnm></au>
    <au><snm>Eggermond</snm><fnm>MA</fnm></au>
    <au><snm>Erath</snm><fnm>A</fnm></au>
  </aug>
  <source>Transportation Research Part C: Emerging Technologies</source>
  <publisher>Elsevier</publisher>
  <pubdate>2021</pubdate>
  <volume>133</volume>
  <fpage>103435</fpage>
</bibl>

<bibl id="B30">
  <title><p>Urban attractors: Discovering patterns in regions of attraction in
  cities</p></title>
  <aug>
    <au><snm>Alhazzani</snm><fnm>M</fnm></au>
    <au><snm>Alhasoun</snm><fnm>F</fnm></au>
    <au><snm>Alawwad</snm><fnm>Z</fnm></au>
    <au><snm>Gonz{\'a}lez</snm><fnm>MC</fnm></au>
  </aug>
  <source>Plos one</source>
  <publisher>Public Library of Science San Francisco, CA USA</publisher>
  <pubdate>2021</pubdate>
  <volume>16</volume>
  <issue>4</issue>
  <fpage>e0250204</fpage>
</bibl>

<bibl id="B31">
  <title><p>Google street view: Capturing the world at street level</p></title>
  <aug>
    <au><snm>Anguelov</snm><fnm>D</fnm></au>
    <au><snm>Dulong</snm><fnm>C</fnm></au>
    <au><snm>Filip</snm><fnm>D</fnm></au>
    <au><snm>Frueh</snm><fnm>C</fnm></au>
    <au><snm>Lafon</snm><fnm>S</fnm></au>
    <au><snm>Lyon</snm><fnm>R</fnm></au>
    <au><snm>Ogale</snm><fnm>A</fnm></au>
    <au><snm>Vincent</snm><fnm>L</fnm></au>
    <au><snm>Weaver</snm><fnm>J</fnm></au>
  </aug>
  <source>Computer</source>
  <publisher>IEEE</publisher>
  <pubdate>2010</pubdate>
  <volume>43</volume>
  <issue>6</issue>
  <fpage>32</fpage>
  <lpage>-38</lpage>
</bibl>

<bibl id="B32">
  <title><p>{Mapillary - Street-level imagery, powered by collaboration and
  computer vision }</p></title>
  <aug>
    <au><cnm>{Mapillary contributors}</cnm></au>
  </aug>
  <source>\url{ https://www.mapillary.com/app }</source>
  <pubdate>2019</pubdate>
</bibl>

<bibl id="B33">
  <title><p>Attention in dichotic listening: Affective cues and the influence
  of instructions</p></title>
  <aug>
    <au><snm>Moray</snm><fnm>N</fnm></au>
  </aug>
  <source>Quarterly journal of experimental psychology</source>
  <publisher>Taylor \& Francis</publisher>
  <pubdate>1959</pubdate>
  <volume>11</volume>
  <issue>1</issue>
  <fpage>56</fpage>
  <lpage>-60</lpage>
</bibl>

<bibl id="B34">
  <title><p>Attention and effort</p></title>
  <aug>
    <au><snm>Kahneman</snm><fnm>D</fnm></au>
  </aug>
  <publisher>Citeseer</publisher>
  <pubdate>1973</pubdate>
  <volume>1063</volume>
</bibl>

<bibl id="B35">
  <title><p>The capacity of visual short-term memory is set both by visual
  information load and by number of objects</p></title>
  <aug>
    <au><snm>Alvarez</snm><fnm>GA</fnm></au>
    <au><snm>Cavanagh</snm><fnm>P</fnm></au>
  </aug>
  <source>Psychological science</source>
  <publisher>SAGE Publications Sage CA: Los Angeles, CA</publisher>
  <pubdate>2004</pubdate>
  <volume>15</volume>
  <issue>2</issue>
  <fpage>106</fpage>
  <lpage>-111</lpage>
</bibl>

<bibl id="B36">
  <title><p>The development of attention to simple and complex visual stimuli
  in infants: Behavioral and psychophysiological measures</p></title>
  <aug>
    <au><snm>Richards</snm><fnm>JE</fnm></au>
  </aug>
  <source>Developmental Review</source>
  <publisher>Elsevier</publisher>
  <pubdate>2010</pubdate>
  <volume>30</volume>
  <issue>2</issue>
  <fpage>203</fpage>
  <lpage>-219</lpage>
</bibl>

<bibl id="B37">
  <title><p>Predicting Driver Self-Reported Stress by Analyzing the Road
  Scene</p></title>
  <aug>
    <au><snm>Bustos</snm><fnm>C</fnm></au>
    <au><snm>Elhaouij</snm><fnm>N</fnm></au>
    <au><snm>Sole Ribalta</snm><fnm>A</fnm></au>
    <au><snm>Borge Holthoefer</snm><fnm>J</fnm></au>
    <au><snm>Lapedriza</snm><fnm>A</fnm></au>
    <au><snm>Picard</snm><fnm>R</fnm></au>
  </aug>
  <source>2021 9th International Conference on Affective Computing and
  Intelligent Interaction (ACII)</source>
  <pubdate>2021</pubdate>
  <fpage>xxx</fpage>
  <lpage>-yyy</lpage>
</bibl>

<bibl id="B38">
  <title><p>Assessing bikeability with street view imagery and computer
  vision</p></title>
  <aug>
    <au><snm>Ito</snm><fnm>K</fnm></au>
    <au><snm>Biljecki</snm><fnm>F</fnm></au>
  </aug>
  <source>arXiv preprint arXiv:2105.08499</source>
  <pubdate>2021</pubdate>
</bibl>

<bibl id="B39">
  <title><p>Using Google Street View to investigate the association between
  street greenery and physical activity</p></title>
  <aug>
    <au><snm>Lu</snm><fnm>Y</fnm></au>
  </aug>
  <source>Landscape and Urban Planning</source>
  <publisher>Elsevier</publisher>
  <pubdate>2019</pubdate>
  <volume>191</volume>
  <fpage>103435</fpage>
</bibl>

<bibl id="B40">
  <title><p>Google street view image of a house predicts car accident risk of
  its resident</p></title>
  <aug>
    <au><snm>Kita</snm><fnm>K</fnm></au>
    <au><snm>Kidzi{\'n}ski</snm><fnm>{\L}</fnm></au>
  </aug>
  <source>arXiv preprint arXiv:1904.05270</source>
  <pubdate>2019</pubdate>
</bibl>

<bibl id="B41">
  <title><p>Streetscore-predicting the perceived safety of one million
  streetscapes</p></title>
  <aug>
    <au><snm>Naik</snm><fnm>N</fnm></au>
    <au><snm>Philipoom</snm><fnm>J</fnm></au>
    <au><snm>Raskar</snm><fnm>R</fnm></au>
    <au><snm>Hidalgo</snm><fnm>C</fnm></au>
  </aug>
  <source>Proceedings of the IEEE Conference on Computer Vision and Pattern
  Recognition Workshops</source>
  <pubdate>2014</pubdate>
  <fpage>779</fpage>
  <lpage>-785</lpage>
</bibl>

<bibl id="B42">
  <title><p>Using deep learning and Google Street View to estimate the
  demographic makeup of neighborhoods across the United States</p></title>
  <aug>
    <au><snm>Gebru</snm><fnm>T</fnm></au>
    <au><snm>Krause</snm><fnm>J</fnm></au>
    <au><snm>Wang</snm><fnm>Y</fnm></au>
    <au><snm>Chen</snm><fnm>D</fnm></au>
    <au><snm>Deng</snm><fnm>J</fnm></au>
    <au><snm>Aiden</snm><fnm>EL</fnm></au>
    <au><snm>Fei Fei</snm><fnm>L</fnm></au>
  </aug>
  <source>Proceedings of the National Academy of Sciences</source>
  <publisher>National Acad Sciences</publisher>
  <pubdate>2017</pubdate>
  <volume>114</volume>
  <issue>50</issue>
  <fpage>13108</fpage>
  <lpage>-13113</lpage>
</bibl>

<bibl id="B43">
  <title><p>Towards Better Driver Safety: Empowering Personal Navigation
  Technologies with Road Safety Awareness</p></title>
  <aug>
    <au><snm>Xu</snm><fnm>R</fnm></au>
    <au><snm>Lin</snm><fnm>AY</fnm></au>
    <au><snm>Zhang</snm><fnm>S</fnm></au>
    <au><snm>Xiong</snm><fnm>P</fnm></au>
    <au><snm>Hecht</snm><fnm>B</fnm></au>
  </aug>
  <source>arXiv preprint arXiv:2006.03196</source>
  <pubdate>2020</pubdate>
</bibl>

<bibl id="B44">
  <title><p>Planning for sustainable Open Streets in pandemic
  cities</p></title>
  <aug>
    <au><snm>Rhoads</snm><fnm>D</fnm></au>
    <au><snm>Sol{\'e} Ribalta</snm><fnm>A</fnm></au>
    <au><snm>Gonz{\'a}lez</snm><fnm>MC</fnm></au>
    <au><snm>Borge Holthoefer</snm><fnm>J</fnm></au>
  </aug>
  <source>arXiv preprint arXiv:2009.12548</source>
  <pubdate>2020</pubdate>
</bibl>

<bibl id="B45">
  <title><p>Personalized routing for multitudes in smart cities</p></title>
  <aug>
    <au><snm>De Domenico</snm><fnm>M</fnm></au>
    <au><snm>Lima</snm><fnm>A</fnm></au>
    <au><snm>Gonz{\'a}lez</snm><fnm>MC</fnm></au>
    <au><snm>Arenas</snm><fnm>A</fnm></au>
  </aug>
  <source>EPJ Data Science</source>
  <publisher>SpringerOpen</publisher>
  <pubdate>2015</pubdate>
  <volume>4</volume>
  <issue>1</issue>
  <fpage>1</fpage>
  <lpage>-11</lpage>
</bibl>

<bibl id="B46">
  <title><p>Open Data BCN</p></title>
  <aug>
    <au><cnm>{Ajuntament de Barcelona}</cnm></au>
  </aug>
  <source>\url{https://opendata-ajuntament.barcelona.cat/en/}</source>
  <pubdate>2019</pubdate>
  <note>Accessed: 2019-04-20</note>
</bibl>

<bibl id="B47">
  <title><p>{Planet dump retrieved from https://planet.osm.org }</p></title>
  <aug>
    <au><cnm>{OpenStreetMap contributors}</cnm></au>
  </aug>
  <source>\url{ https://www.openstreetmap.org }</source>
  <pubdate>2017</pubdate>
</bibl>

<bibl id="B48">
  <title><p>Deep residual learning for image recognition</p></title>
  <aug>
    <au><snm>He</snm><fnm>K</fnm></au>
    <au><snm>Zhang</snm><fnm>X</fnm></au>
    <au><snm>Ren</snm><fnm>S</fnm></au>
    <au><snm>Sun</snm><fnm>J</fnm></au>
  </aug>
  <source>Proceedings of the IEEE Conference on Computer Vision and Pattern
  Recognition</source>
  <pubdate>2016</pubdate>
  <fpage>770</fpage>
  <lpage>-778</lpage>
</bibl>

<bibl id="B49">
  <title><p>A sustainable strategy for Open Streets in (post) pandemic
  cities</p></title>
  <aug>
    <au><snm>Rhoads</snm><fnm>D</fnm></au>
    <au><snm>Sol{\'e} Ribalta</snm><fnm>A</fnm></au>
    <au><snm>Gonz{\'a}lez</snm><fnm>MC</fnm></au>
    <au><snm>Borge Holthoefer</snm><fnm>J</fnm></au>
  </aug>
  <source>Communications Physics</source>
  <publisher>Nature Publishing Group</publisher>
  <pubdate>2021</pubdate>
  <volume>4</volume>
  <issue>1</issue>
  <fpage>1</fpage>
  <lpage>-12</lpage>
</bibl>

<bibl id="B50">
  <title><p>Limits of predictability in commuting flows in the absence of data
  for calibration</p></title>
  <aug>
    <au><snm>Yang</snm><fnm>Y</fnm></au>
    <au><snm>Herrera</snm><fnm>C</fnm></au>
    <au><snm>Eagle</snm><fnm>N</fnm></au>
    <au><snm>Gonz{\'a}lez</snm><fnm>MC</fnm></au>
  </aug>
  <source>Scientific reports</source>
  <publisher>Nature Publishing Group</publisher>
  <pubdate>2014</pubdate>
  <volume>4</volume>
  <issue>1</issue>
  <fpage>1</fpage>
  <lpage>-9</lpage>
</bibl>

<bibl id="B51">
  <title><p>Participatory Cultural Mapping Based on Collective Behavior Data in
  Location-Based Social Networks</p></title>
  <aug>
    <au><snm>Yang</snm><fnm>D</fnm></au>
    <au><snm>Zhang</snm><fnm>D</fnm></au>
    <au><snm>Qu</snm><fnm>B</fnm></au>
  </aug>
  <source>ACM Transactions on Intelligent Systems and Technology
  (TIST)</source>
  <publisher>ACM</publisher>
  <pubdate>2016</pubdate>
  <volume>7</volume>
  <issue>3</issue>
  <fpage>30</fpage>
</bibl>

<bibl id="B52">
  <title><p>NationTelescope: Monitoring and visualizing large-scale collective
  behavior in LBSNs</p></title>
  <aug>
    <au><snm>Yang</snm><fnm>D</fnm></au>
    <au><snm>Zhang</snm><fnm>D</fnm></au>
    <au><snm>Chen</snm><fnm>L</fnm></au>
    <au><snm>Qu</snm><fnm>B</fnm></au>
  </aug>
  <source>Journal of Network and Computer Applications</source>
  <publisher>Elsevier</publisher>
  <pubdate>2015</pubdate>
  <volume>55</volume>
  <fpage>170</fpage>
  <lpage>-180</lpage>
</bibl>

<bibl id="B53">
  <title><p>Decongestion of urban areas with hotspot pricing</p></title>
  <aug>
    <au><snm>Sol{\'e} Ribalta</snm><fnm>A</fnm></au>
    <au><snm>G{\'o}mez</snm><fnm>S</fnm></au>
    <au><snm>Arenas</snm><fnm>A</fnm></au>
  </aug>
  <source>Networks and Spatial Economics</source>
  <publisher>Springer</publisher>
  <pubdate>2018</pubdate>
  <volume>18</volume>
  <issue>1</issue>
  <fpage>33</fpage>
  <lpage>-50</lpage>
</bibl>

<bibl id="B54">
  <title><p>Spatio-temporal correlations of betweenness centrality and traffic
  metrics</p></title>
  <aug>
    <au><snm>Henry</snm><fnm>E</fnm></au>
    <au><snm>Bonnetain</snm><fnm>L</fnm></au>
    <au><snm>Furno</snm><fnm>A</fnm></au>
    <au><snm>El Faouzi</snm><fnm>NE</fnm></au>
    <au><snm>Zimeo</snm><fnm>E</fnm></au>
  </aug>
  <source>2019 6th International Conference on Models and Technologies for
  Intelligent Transportation Systems (MT-ITS)</source>
  <pubdate>2019</pubdate>
  <fpage>1</fpage>
  <lpage>-10</lpage>
</bibl>

<bibl id="B55">
  <title><p>Augmented betweenness centrality for mobility prediction in
  transportation networks</p></title>
  <aug>
    <au><snm>Altshuler</snm><fnm>Y</fnm></au>
    <au><snm>Puzis</snm><fnm>R</fnm></au>
    <au><snm>Elovici</snm><fnm>Y</fnm></au>
    <au><snm>Bekhor</snm><fnm>S</fnm></au>
    <au><snm>Pentland</snm><fnm>AS</fnm></au>
  </aug>
  <source>International Workshop on Finding Patterns of Human Behaviors in
  NEtworks and MObility Data, NEMO11</source>
  <pubdate>2011</pubdate>
</bibl>

<bibl id="B56">
  <title><p>La Guardia Urbana comprometida con el objetivo Vision
  Cero</p></title>
  <source>La Guardia Urbana comprometida con el objetivo Visión Cero | Guardia
  Urbana de Barcelona | Ajuntament de Barcelona</source>
  <pubdate>2017</pubdate>
</bibl>

<bibl id="B57">
  <title><p>Changing the urban design of cities for health: The superblock
  model</p></title>
  <aug>
    <au><snm>Mueller</snm><fnm>N</fnm></au>
    <au><snm>Rojas Rueda</snm><fnm>D</fnm></au>
    <au><snm>Khreis</snm><fnm>H</fnm></au>
    <au><snm>Cirach</snm><fnm>M</fnm></au>
    <au><snm>Andr{\'e}s</snm><fnm>D</fnm></au>
    <au><snm>Ballester</snm><fnm>J</fnm></au>
    <au><snm>Bartoll</snm><fnm>X</fnm></au>
    <au><snm>Daher</snm><fnm>C</fnm></au>
    <au><snm>Deluca</snm><fnm>A</fnm></au>
    <au><snm>Echave</snm><fnm>C</fnm></au>
    <au><cnm>others</cnm></au>
  </aug>
  <source>Environment international</source>
  <publisher>Elsevier</publisher>
  <pubdate>2020</pubdate>
  <volume>134</volume>
  <fpage>105132</fpage>
</bibl>

<bibl id="B58">
  <title><p>Tactical urbanism: Towards an evolutionary cities’
  approach?</p></title>
  <aug>
    <au><snm>Silva</snm><fnm>P</fnm></au>
  </aug>
  <source>Environment and Planning B: Planning and design</source>
  <publisher>SAGE Publications Sage UK: London, England</publisher>
  <pubdate>2016</pubdate>
  <volume>43</volume>
  <issue>6</issue>
  <fpage>1040</fpage>
  <lpage>-1051</lpage>
</bibl>

<bibl id="B59">
  <title><p>Networks: An Introduction</p></title>
  <aug>
    <au><snm>Newman</snm><fnm>M.</fnm></au>
  </aug>
  <publisher>OUP Oxford</publisher>
  <pubdate>2009</pubdate>
</bibl>

<bibl id="B60">
  <title><p>A set of measures of centrality based on betweenness</p></title>
  <aug>
    <au><snm>Freeman</snm><fnm>LC</fnm></au>
  </aug>
  <source>Sociometry</source>
  <publisher>JSTOR</publisher>
  <pubdate>1977</pubdate>
  <fpage>35</fpage>
  <lpage>-41</lpage>
</bibl>

<bibl id="B61">
  <title><p>A Road Segment Prioritization Approach for Cycling
  Infrastructure</p></title>
  <aug>
    <au><snm>Mahfouz</snm><fnm>H</fnm></au>
    <au><snm>Arcaute</snm><fnm>E</fnm></au>
    <au><snm>Lovelace</snm><fnm>R</fnm></au>
  </aug>
  <source>arXiv preprint arXiv:2105.03712</source>
  <pubdate>2021</pubdate>
</bibl>

<bibl id="B62">
  <title><p>In-place activated batchnorm for memory-optimized training of
  dnns</p></title>
  <aug>
    <au><snm>Bulo</snm><fnm>SR</fnm></au>
    <au><snm>Porzi</snm><fnm>L</fnm></au>
    <au><snm>Kontschieder</snm><fnm>P</fnm></au>
  </aug>
  <source>Proceedings of the IEEE Conference on Computer Vision and Pattern
  Recognition</source>
  <pubdate>2018</pubdate>
  <fpage>5639</fpage>
  <lpage>-5647</lpage>
</bibl>

<bibl id="B63">
  <title><p>The mapillary vistas dataset for semantic understanding of street
  scenes</p></title>
  <aug>
    <au><snm>Neuhold</snm><fnm>G</fnm></au>
    <au><snm>Ollmann</snm><fnm>T</fnm></au>
    <au><snm>Rota Bulo</snm><fnm>S</fnm></au>
    <au><snm>Kontschieder</snm><fnm>P</fnm></au>
  </aug>
  <source>Proceedings of the IEEE international conference on computer
  vision</source>
  <pubdate>2017</pubdate>
  <fpage>4990</fpage>
  <lpage>-4999</lpage>
</bibl>

</refgrp>
} 








%

\end{document}